\documentclass[10pt,twocolumn,letterpaper]{article}

\usepackage{cvpr}              % To produce the CAMERA-READY version
\definecolor{cvprblue}{rgb}{0.21,0.49,0.74}
\usepackage[pagebackref,breaklinks,colorlinks,allcolors=cvprblue]{hyperref}

\def\paperID{16237} % *** Enter the Paper ID here
\def\confName{CVPR}
\def\confYear{2026}
\usepackage{algorithm}
\usepackage[noend]{algorithmic}
\usepackage{multirow}
\usepackage{xcolor}
\usepackage{colortbl, booktabs}
\usepackage{amsmath}
\usepackage{pifont}
\usepackage{makecell}
\usepackage{color,xcolor}
\definecolor{sgy}{rgb}{0,0,0}

\title{SODA: Sensitivity-Oriented Dynamic Acceleration for Diffusion Transformer}

\author{Tong Shao, Yusen Fu, Guoying Sun, Jingde Kong, Zhuotao Tian\thanks{Corresponding authors.}, Jingyong Su\footnotemark[1]\\
Harbin Institute of Technology, Shenzhen\\
{\tt\small \{shaotong, 220110516, sunguoying, 2023113058\}@stu.hit.edu.cn},\\
{\tt\small \{tianzhuotao, sujingyong\}@hit.edu.cn}
}

\newcommand{\mypara}[1]{\noindent\textbf{#1}}

\begin{document}
\maketitle
\begin{abstract}
Diffusion Transformers have become a dominant paradigm in visual generation, yet their low inference efficiency remains a key bottleneck hindering further advancement. Among common training-free techniques, caching offers high acceleration efficiency but often compromises fidelity, whereas pruning shows the opposite trade-off. Integrating caching with pruning achieves a balance between acceleration and generation quality. However, existing methods typically employ fixed and heuristic schemes to configure caching and pruning strategies. While they roughly follow the overall sensitivity trend of generation models to acceleration, they fail to capture fine-grained and complex variations, inevitably skipping highly sensitive computations and leading to quality degradation. Furthermore, such manually designed strategies exhibit poor generalization. To address these issues, we propose SODA, a Sensitivity-Oriented Dynamic Acceleration method that adaptively performs caching and pruning based on fine-grained sensitivity. SODA builds an offline sensitivity error modeling framework across timesteps, layers, and modules to capture the sensitivity to different acceleration operations. The cache intervals are optimized via dynamic programming with sensitivity error as the cost function, minimizing the impact of caching on model sensitivity. During pruning and cache reuse, SODA adaptively determines the pruning timing and rate to preserve computations of highly sensitive tokens, significantly enhancing generation fidelity. Extensive experiments on DiT-XL/2, PixArt-$\alpha$, and OpenSora demonstrate that SODA achieves state-of-the-art generation fidelity under controllable acceleration ratios. Our code is released publicly at: https://github.com/leaves162/SODA.
\end{abstract}    
\section{Introduction}
\label{sec:intro}
Diffusion models~\cite{ho2020denoising,rombach2022high} have demonstrated substantial advancements in visual generation, with notable success in image~\cite{chen2023pixart} and video~\cite{zheng2024open} generation.
With the introduction of Diffusion Transformer (DiT)~\cite{peebles2023scalable}, the generation quality has been further significantly improved, promoting the numerous applications such as image editing~\cite{zhang2023adding} and medical image analysis~\cite{guo2025survey,sun2025braincognizer}.
Despite its impressive performance, the low inference efficiency caused by repeated sampling timesteps and transformer blocks poses a significant challenge to real-world deployment, especially in resource-constrained or latency-sensitive scenarios.

\begin{figure*}[t]
\centering
\includegraphics[width=1.0\textwidth]{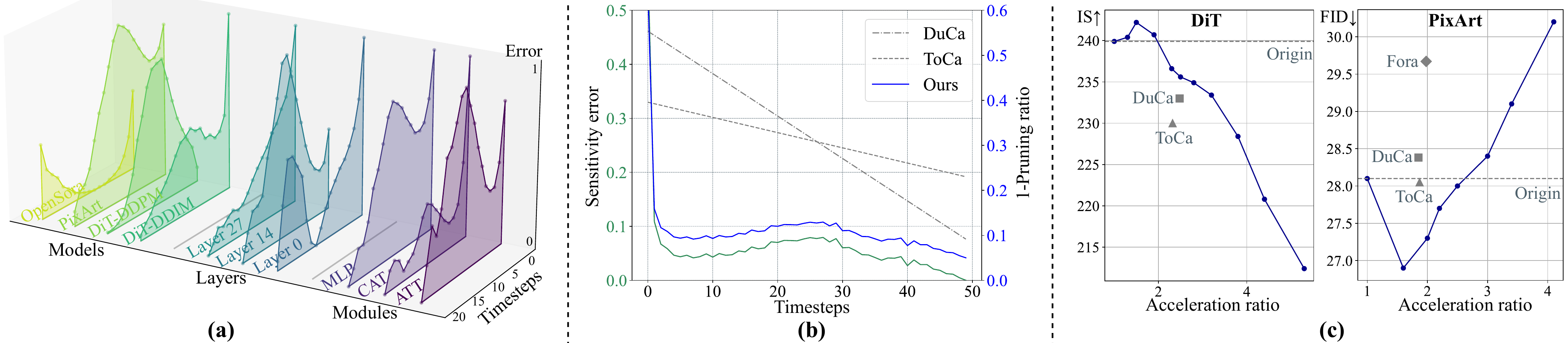} % Reduce the figure size so that it is slightly narrower than the column.
\caption{\textbf{Our SODA enables sensitivity awareness and adaptive acceleration strategy decision.} \textbf{(a).} Examples of model-internal sensitivity to acceleration. The sensitivity of different timesteps, layers, and modules to acceleration is highly complex and dynamic. \textbf{(b).} Correlation between acceleration strategies and model sensitivity. Heuristic methods such as DuCa~\cite{zou2024accelerating1} fail to be consistent with the complex sensitivity. \textbf{(c).} Acceleration versus generation quality. Our SODA significantly outperforms the baselines by alleviating quality degradation under acceleration.
}
\label{fig:head}
\end{figure*}

Training-based acceleration methods such as distillation~\cite{yin2024one,salimans2022progressive} or finetuning~\cite{zhu2025dig,yu2024image,fang2025attend} incur substantial computational costs. Therefore, recent research has increasingly favored training-free acceleration, such as caching~\cite{ma2024deepcache,liu2025timestep,saghatchian2025cached,qiu2025accelerating} and pruning~\cite{cheng2025cat,saghatchian2025cached,sun2024asymrnr}.
Owing to the similarity between adjacent timesteps and the token-wise redundancy in the diffusion, current pipelines~\cite{selvaraju2024fora,zou2024accelerating,zou2024accelerating1} performs full computation at specific timesteps (\textit{anchor timesteps}) and caches intermediate hidden states. During subsequent timesteps (\textit{pruning timesteps}) within the interval between two fully computed anchor timesteps (\textit{cache interval}), redundant tokens with high similarity are pruned and substituted with cache.
The integration of caching and pruning combines the high acceleration of caching with the structural flexibility of pruning.
It not only enables fine-grained token omission and corrects the caching error, but also offers higher efficiency relative to pure pruning.

Although this pipeline attains a balance between efficiency and quality, it heavily relies on fixed~\cite{bolya2023token,zou2025exposure} or heuristic~\cite{lv2024fastercache,zou2024accelerating} configurations.
For example, FasterDiffusion~\cite{li2023faster} employs a fixed set of anchor timesteps. ToCa~\cite{zou2024accelerating} and DuCa~\cite{zou2024accelerating1} design a manually crafted pruning strategy, where the pruning ratio is increased during the denoising.
% However, such approaches only approximate coarse sensitivity trends of generation models to acceleration, \textit{failing to capture fine-grained sensitivity variations across timesteps, layers, or modules, thus inevitably skip highly sensitive modules, leading to degradation in generation fidelity.}
{\color{sgy}However, such approaches merely capture coarse sensitivity trends of generation models under acceleration, \textit{neglecting fine-grained variations across timesteps, layers, or modules, and consequently skipping highly sensitive components, which ultimately degrades generation fidelity.}}
Moreover, heuristic methods rely on experience‑based design, \textit{which hinders cross-model generalization.}

\textbf{Key observations.}
To illustrate the internal acceleration sensitivity of generation models, we perform fine-grained sensitivity modeling inspired by~\cite{cem}.
As shown in Fig.~\ref{fig:head}(a), the internal sensitivity is highly complex and dynamic.
Coarse experience-based or fixed acceleration strategies in Fig.~\ref{fig:head}(b) are unable to perceive these variations, 
{\color{sgy}inevitably skipping highly sensitive computations.}
Following this observation, a critical question arises:
\textit{How can we enable acceleration strategies to perceive the intricate sensitivity patterns without sacrificing efficiency?}

\textbf{Our solution.}
To tackle this issue, we propose \textbf{SODA}, a \textbf{S}ensitivity-\textbf{O}riented \textbf{D}ynamic \textbf{A}cceleration framework that adaptively integrates sensitivity perception with acceleration strategies, unifying the decision-making of caching and pruning.
SODA quantifies sensitivity to caching and pruning across timesteps, layers and modules by measuring the error between accelerated and GT features. The modeling is conducted offline to avoid affecting inference efficiency.
Leveraging the sensitivity errors as dynamic costs, SODA integrates dynamic programming to optimize the combination of caching intervals, ensuring minimal cumulative error.
Moreover, SODA adaptively determines the pruning intervention timing and rate to skip insensitive tokens and replace them with cache by comparing the sensitivity errors of caching and pruning, improving generation fidelity.

% associated with caching and pruning, we automatically select the operation that yields the smaller error, enabling adaptive acceleration strategy decision.
% Moreover, SODA integrates dynamic programming to optimize the combination of caching intervals, ensuring minimal cumulative sensitivity error.

% Compared to other methods, EAA precisely captures acceleration sensitivity across denoising, retains essential computations to preserve output fidelity, and achieves generalized adaptive acceleration across different generation models.
{\color{sgy}Compared to other methods, SODA precisely captures acceleration sensitivity across denoising, retains essential computations to preserve output fidelity, and achieves generalized adaptive acceleration across different generation models.}
Extensive experiments across three tasks demonstrate that SODA achieves a better balance between acceleration and generation quality.
In Fig.~\ref{fig:head}(c), SODA allows dynamic control over the acceleration budget, and even improves the original model performance under 1.5× acceleration.
% This highlights the effectiveness of our method with fine-grained error modeling and adaptive acceleration to simultaneously improve inference efficiency and exploit the model's performance capacity.
To summarize, our contributions are as follows:
\begin{itemize}
    \item We propose SODA, a sensitivity-oriented dynamic acceleration method that adaptively achieves caching and pruning decisions based on fine-grained sensitivity. It requires no empirical heuristics or manual design and exhibits strong cross-model generalization capability.
    \item We integrate dynamic programming to derive a globally optimal cache‑interval combination strategy that ensures minimal cumulative sensitivity error and introduces no additional overhead.
    \item We unify the caching and pruning decisions based on sensitivity error, adaptively determining the pruning timing and rate to skip only insensitive tokens, thereby improving the fidelity of accelerated generation.
    \item Extensive experiments on DiT-XL/2, PixArt-$\alpha$, and OpenSora demonstrate that SODA can achieve superior generation quality under flexible acceleration ratios.
\end{itemize}
\section{Preliminary}
\label{sec:Preliminary}
\textbf{Diffusion Transformer.}
Diffusion models~\cite{ho2020denoising,rombach2022high} synthesize visual content from noise.
When reversing from timestep $T$ to $1$, $\boldsymbol{x_{t}}$ to $\boldsymbol{x_{t-1}}$ can be modeled as:
\begin{equation}
    \boldsymbol{x_{t-1}} = \frac{1}{\sqrt{a_{t}}}(\boldsymbol{x_{t}} - \frac{1-a_{t}}{\sqrt{1-\bar a_{t}}}\boldsymbol{\boldsymbol{\epsilon}_{\theta}}(\boldsymbol{x_{t}},t))+\sigma_{t}\boldsymbol{z},
\end{equation}
where $a_{t}$, $\bar a_{t}$, $\sigma_{t}$ are constants related to $t$, $\boldsymbol{{\epsilon}_{\theta}}(\boldsymbol{x_{t}},t)$ is the noise estimate parameterized by $\boldsymbol{\theta}$, and $\boldsymbol{z} \sim \mathcal{N}(\boldsymbol{0}, \boldsymbol{I})$.

Diffusion Transformer~\cite{peebles2023scalable} integrates the transformer~\cite{vaswani2017attention} into the diffusion process, enabling improved controllability and higher generation quality compared with U-Net~\cite{ronneberger2015u}.
Although DiT has achieved substantial progress, its slow inference process remains a bottleneck, limiting its broader adoption and real-world applicability.

\begin{figure*}[t]
\centering
\includegraphics[width=1.0\textwidth]{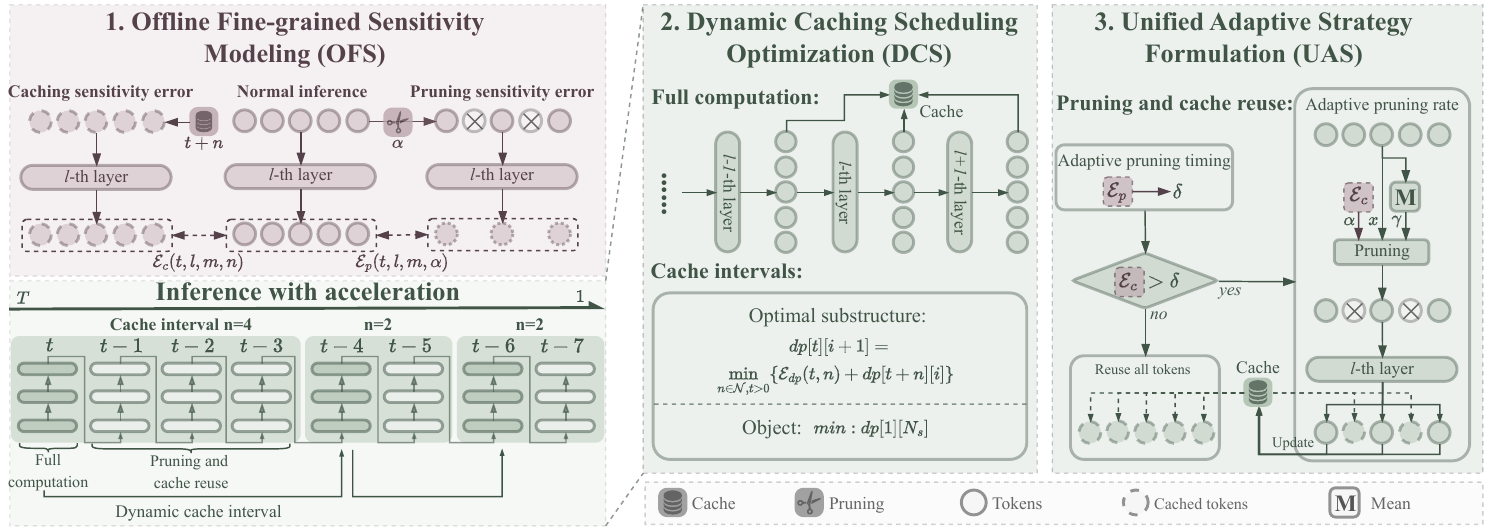} % Reduce the figure size so that it is slightly narrower than the column.
\caption{\textbf{Model architecture of SODA.} We propose SODA, a Sensitivity-Oriented Dynamic Acceleration method. \textbf{(1).} Offline Fine-grained Sensitivity Modeling (OFS): Defining error to measure the fine-grained sensitivity of timesteps, layers and modules before inference. \textbf{(2).} Dynamic Caching Scheduling Optimization (DCS): Employing dynamic programming to identify the optimal combination of cache intervals that yields the minimal sensitivity impact. \textbf{(3).} Unified Adaptive Strategy Formulation (UAS): Achieving adaptive scheduling for pruning timing and rate guided by sensitivity errors when pruning.}
% We proposes Prior Error Estimation to analyze the complexity of acceleration sensitivity. Based on estimated error, EAA proposes Error-Aware Dynamic Caching to select the optimal set of anchor timesteps and cache intervals. Furthermore, Error-Aware Adaptive Pruning is designed at pruning timesteps to further reduce the caching error.
\label{fig:model}
\end{figure*}
% 分模块没写

\mypara{Training-free DiT Acceleration.}
Training methods such as distillation~\cite{meng2023distillation,yao2024fasterdit} represent intuitive solutions, however, they incur high computational overhead and limited generalization.
Among training-free methods, caching~\cite{chen2024delta,zhao2024real,liu2024faster,kahatapitiya2024adaptive,aggarwal2025evolutionary} over adjacent timesteps achieves high efficiency, but leads to quality degradation due to skipping critical components.
Inspired by large language model (LLM) acceleration, token reduction~\cite{wang2024attention,cheng2025cat,zhang2024token,ding2025efficient} eliminates tokens to mitigate the complexity of attention. While pruning offers greater flexibility, its efficiency is lower than caching.

The integration of cache and pruning combines the high efficiency of cache with the flexibility of token-level pruning.
% In each cache interval, the first timestep (\textit{anchor timestep}) is fully computed and cached.
% For the subsequent timesteps (\textit{pruning timestep}), a portion of the tokens is computed from scratch, while the remaining tokens are reused from the cache to reduce redundant computation.
However, fixed or heuristic strategies like~\cite{zou2024accelerating,zou2024accelerating1} rely on predefined cache intervals and pruning rates, which fail to adapt to the diverse and complex sensitivity across timesteps, layers, and modules.
To address it, this paper proposes an adaptive acceleration, which adaptively determines the strategy based on fine-grained sensitivity.

Moreover, this pipeline can be integrated with other general acceleration techniques, such as denoising strategies (DDIM~\cite{song2020denoising}), quantization (FP16 to INT8)~\cite{shang2023post,liu2025cachequant}, and CUDA hardware acceleration (FlashAttention)~\cite{dao2022flashattention}. Our SODA is compatible with these techniques in theory.
Refer to the Appendix D for more related works.
\section{Methodology}
\label{sec:Method}
To adapt to the complex internal sensitivity to acceleration, we propose SODA.
As illustrated in Fig.~\ref{fig:model}, it first conducts
\textbf{(1) Offline Fine-grained Sensitivity Modeling (OFS)}: Defining error to measure the fine-grained sensitivity of timesteps, layers and modules before inference.
Then, SODA adopts \textbf{(2) Dynamic Caching Scheduling Optimization (DCS)}: Employing dynamic programming to identify the optimal combination of cache intervals that yields the minimal sensitivity impact.
Finally, when pruning and cache reuse, SODA proposes
\textbf{(3) Unified Adaptive Strategy Formulation (UAS)}: Achieving adaptive scheduling for pruning timing and rate guided by sensitivity errors.

% In this paper, we define the denoising inference process as progressing from timestep $T$ to $1$, and denote $\mathcal{D}(t,l,x)$ as the inference result at a specific timestep $t$ and layer $l$, given input feature $x$.

\subsection{Offline Fine-grained Sensitivity Modeling}
\label{sec:ofs}
To perceive the internal acceleration sensitivity of generative models, we model the sensitivity of different timesteps, layers, and modules to various acceleration operations and efficiencies.
Specifically, we model sensitivity by defining the error between the accelerated and GT features.
For caching, we evaluate each module’s sensitivity error under varying acceleration efficiencies, i.e., different cache intervals (the number of steps between cache updates).
For pruning, we measure the sensitivity error for different pruning ratios, indicating how many tokens are removed.

\textbf{Caching sensitivity error.}
It aims to quantify the error of reusing the previous timestep $t+n$ at current timestep $t$ (Denoising proceeds backward from $T$ to $1$, where cache interval is $n$).
The sensitivity error $\mathcal{E}_c(t,l,m,n)$ is measured as the Cosine distance ($\texttt{Cos}$) between the reused output $\mathcal{D}_{t+n,l,m}(x)$ and the corresponding target output $\mathcal{D}_{t,l,m}(x)$:
\begin{equation}
    \mathcal{E}_c(t,l,m,n)=1-\texttt{Cos}(\mathcal{D}_{t+n,l,m}(x),\mathcal{D}_{t,l,m}(x)),
\end{equation}
where $\mathcal{D}$ is the diffusion at timestep $t$, layer $l$ and module $m$, $x$ is input and $n$ denotes cache interval. Taking DiT-XL/2 with DDIM as an example, $t\in[1,50],l\in[1,28],m\in\{\text{ATT},\text{MLP}\}$, and $n\in[1,9]$ to cover as wide a range of acceleration ratios as possible during modeling.

\textbf{Pruning sensitivity error.}
It is computed analogously and serves to evaluate the pruning error for each module:
\begin{equation}
    \mathcal{E}_p(t,l,m,\alpha)=1-\texttt{Cos}(\mathcal{D}_{t,l,m}(x'),\mathcal{D}_{t,l,m}(x)),
\end{equation}
where $x'$ denotes the feature inputs after pruning operation $x'=\mathcal{P}_{\alpha,\gamma}(x)$, while $\alpha$ and $\gamma$ correspond to the pruning ratio and the region, respectively, $\alpha \in [0.1, 0.9]$ with a step size of 0.1 for capturing sensitivity trends under all acceleration efficiencies, and the position $\gamma$ is randomized during modeling, mitigating the influence of the generated content.

\textbf{Offline modeling.}
The modeling captures fine-grained sensitivity across timesteps, layers and modules, preventing fidelity degradation caused by skipping highly sensitive modules.
However, it inevitably needs additional computational overhead, hindering acceleration efficiency.
Moreover, the sensitivity is model-specific, making repeated modeling during inference redundant and wasteful.

Therefore, SODA performs the above modeling process offline by generating multiple random contents and averaging them, storing the sensitivity errors as model-specific priors.
It makes the sensitivity error model-specific and independent of the generated content, so each generative model only needs to be modeled once for permanent use.

Details regarding the consistency between the prior and formal inference error distributions, the number of random generations, and the offline computation cost are provided in Sec.~\ref{sec:offline}.
More sensitivity analysis and visualizations are provided in Appendix A.1.

\subsection{Standard Caching-Pruning Pipeline}
Assume that a full computation is performed at step $t + n$, and the cache interval is set to $n$.
During steps from $t+n-1$ to $t+1$, pruning is applied to skip insensitive tokens and replace them with cache, while sensitive tokens are computed and their corresponding caches are updated.
The process is repeated after a complete computation at step $t$, as depicted in the lower‑left portion of Fig.~\ref{fig:model}.

In previous methods~\cite{zou2024accelerating,zou2024accelerating1}, important parameters including \textbf{1)} the cache interval $n$, \textbf{2)} the pruning intervention timing \textbf{3)} pruning rate $\alpha$ are manually set and cannot adapt to dynamic changes of sensitivity errors in denoising, which often leads to missing computation of important tokens and fidelity degradation.
To address the above issues, SODA leverages sensitivity errors to develop an adaptive acceleration strategy that improves fidelity.
Sec.~\ref{sec:dcs} solves \textbf{1)} and Sec.~\ref{sec:uas} solves \textbf{2)} and \textbf{3)}.

\subsection{Dynamic Caching Scheduling Optimization}
\label{sec:dcs}
When the effectiveness of a caching strategy can be evaluated in terms of sensitivity error, the cache-interval combination inherently possesses an optimal substructure: the minimum cumulative error at each caching builds upon the optimal result of the preceding operation.
So SODA employs dynamic programming, inspired by~\cite{cem}, as shown in Alg.~\ref{alg:dp}, to optimize the combination of cache intervals.

\textbf{Problem setup.}
Given a budget specified by the number of cache intervals (caching times) $N_s$, let $n$ denote the current cache interval, $t+n$ and $t$ represent the timesteps with full computation with the same setting as above. 

\textbf{Optimal substructure.}
The dynamic programming array $dp[t][i]$ denotes the minimum cumulative error from $T$ to $t$ performing $i$ cache intervals. This array captures the optimal substructure the set of candidate cache intervals $\mathcal{N}$ from when $t+n\rightarrow t$:
\begin{equation}
    dp[t][i+1]=\mathop{min}\limits_{n\in \mathcal{N},t>0}\{\mathcal{E}_{dp}(t,n)+dp[t+n][i]\},
\end{equation}
where $\mathcal{E}_{dp}(t,n)=\frac{\xi}{L\cdot M}\sum_{l=1}^{L}\sum_{m=1}^{M}\mathcal{E}_{c}(t,l,m,n)$, $\mathcal{E}_{dp}$ is the average of $\mathcal{E}_{c}$ over $l$ and $m$, as this process is independent of the layer and module. $\xi$ is an $n$-dependent weighting coefficient used to amplify the differences between different caching intervals. The final objective is to solve for the minimum $dp[1][N_s]$, which means the caching process from step $T$ to $1$ using $N_s$ times. With $dp[1][N_s]$, a backtracking procedure is employed to recover the positions of the selected timesteps and their cache intervals. Refer to Appendix B.1 for implementation details.

\textbf{Analysis.}
Dynamic programming enables SODA to obtain the optimal caching strategy.
As shown in Fig.~\ref{fig:error_performance}, DCS module effectively reduces the sensitivity error and improves generation fidelity under the same acceleration efficiency.
The algorithm operates on offline‑stored sensitivity errors, given an acceleration budget, it can derive the optimal cache‑interval combination that can be shared across multiple generations with no additional overhead. More details are provided in Appendix A.2.

\begin{algorithm}[t]
\renewcommand{\algorithmicrequire}{\textbf{Input:}}
\renewcommand{\algorithmicensure}{\textbf{Output:}}
\caption{Dynamic Caching Scheduling Optimization}
\label{alg:dp}
\begin{algorithmic}[1]
\REQUIRE Total steps $T$, caching times $N_s$, cache interval candidate set $\mathcal{N}$, sensitivity error here $\mathcal{E}_{dp}(t,n)$
\ENSURE timestep set $\mathcal{A}$ and cache interval set $\mathcal{I}$
\STATE $dp[t][i]\gets \infty$, $dp[T][1]\gets 0$, $prev[t][i]\gets \text{None}$ \\
\FOR{$i = 1$ to $N_s$}
    \FOR{$t = T$ down to $1$}
        % \IF{$dp[t][i] < \infty$}
            \FORALL{$n \in \mathcal{N}$}
                % \STATE $t_{\text{next}} \gets t - n$
                \IF{$t > 0$}
                    % \STATE $cost \gets \mathcal{E}_a(t, n)$
                    \IF{$dp[t][i+1] > dp[t+n][i] + \mathcal{E}_{dp}(t,n)$}
                        \STATE $dp[t][i+1] \gets dp[t+n][i] + \mathcal{E}_{dp}(t,n)$
                        \STATE $prev[t][i+1] \gets (t+n, n)$
                    \ENDIF
                \ENDIF
            \ENDFOR
        % \ENDIF
    \ENDFOR
\ENDFOR
\STATE Backtracking: $\mathcal{A}\gets \{\}$, $\mathcal{I}\gets \{\}$, $t \gets 1, i\gets N_s$
% \STATE Initialize: cache interval set $\mathcal{I}\gets \{\}$,
% \STATE Backtracking start: $t \gets 1, \quad k \gets N_s$
\WHILE{$i > 0$}
    \STATE $(t_{\text{next}}, n) \gets \text{prev}[t][i]$
    \STATE $\mathcal{A} \gets \mathcal{A} \cup \{t\}$, $\mathcal{I} \gets \mathcal{I} \cup \{n\}$
    \STATE $t \gets t_{\text{next}}$, $i \gets i - 1$
\ENDWHILE
% \STATE Reverse $anchors$ and $intervals$
\RETURN $\mathcal{A}$, $\mathcal{I}$
\end{algorithmic}
\end{algorithm}

\subsection{Unified Adaptive Strategy Formulation}
\label{sec:uas}
After performing the full computation at step $t + n$ and determining the cache interval $n$, insensitive tokens from $t + n - 1$ to $t + 1$ are skipped through pruning and replaced with cache to achieve acceleration.
During this process, SODA introduces the UAS module to enable adaptive determination of the pruning timing and pruning rate.

\textbf{Adaptive pruning intervention timing.}
To adaptively determine when to apply pruning, we use sensitivity error to unify the scheduling of pruning and caching.
Conceptually, the integration of caching and pruning aims to preserve a portion of sensitive tokens through pruning, thereby reducing sensitivity error and mitigating fidelity degradation.
Yet, pruning itself incurs errors.
Our key insight for adaptive decision is that pruning should be performed if and only if its error is lower than the caching error under the same efficiency.
This design guarantees that pruning contributes to reducing the overall error. Otherwise, pruning would be meaningless, even introducing unnecessary overhead.

Specifically, we fully compute the features of timestep $t+n$, layer $l$ and module $m$, then store its intermediate activations in cache $\mathcal{C}_{l,m}(\Omega) = \mathcal{D}_{t+n,l,m}(x)$, where $\Omega$ denotes the full set of token position indices.

Taking pruning and cache reuse at timestep $t+1$ as an example,
we use the previously modeled $\mathcal{E}_p$ to dynamically determine the pruning rate $\alpha_{t+1,l,m}$ at each node (it is formally defined in the next section).
Our pruning is applied only when its error is lower than the caching error, meaning that pruning can effectively mitigate acceleration sensitivity.
SODA constructs the pruning decision threshold $\delta$ based on the current pruning rate $\alpha_{t+1,l,m}$ and the pruning error $\mathcal{E}_p$: $\delta_{t+1,l,m}=\mathcal{E}_p(t+1,l,m,\alpha_{t+1,l,m})$.
Pruning is allowed only if its $\delta_{t+1,l,m}$ is smaller than the caching error $\mathcal{E}_c(t,l,m,n)$, otherwise, the cache $\mathcal{C}_{l,m}$ is directly reused to avoid larger error.
The final feature $x'$ achieves an adaptive decision:
\begin{equation}
x'=\left\{
             \begin{array}{lr}
             \mathcal{C}_{l,m}(\Omega),\ if\ \delta_{t+1,l,m}\geq\mathcal{E}_c(t,l,m,n),\\
             \mathcal{P}_{\alpha_{t+1,l,m},\gamma}(x)\cup \mathcal{C}_{l,m}(\gamma),\ else.
             \end{array}
\right.
\end{equation}
If the threshold is satisfied, the result is the complement of the pruned computation under pruning $\mathcal{P}$ and cache $\mathcal{C}$ at the corresponding position indices set $\gamma$,
$\cup$ represents the complement. Here, the positions are determined by TopK selection using the feature mean as the importance metric, which avoids the incompatibility caused by relying on attention weights under Flash Attention.
Details of the pruning can be found in Appendix A.3.

\begin{figure}[t]
\centering
\includegraphics[width=0.47\textwidth]{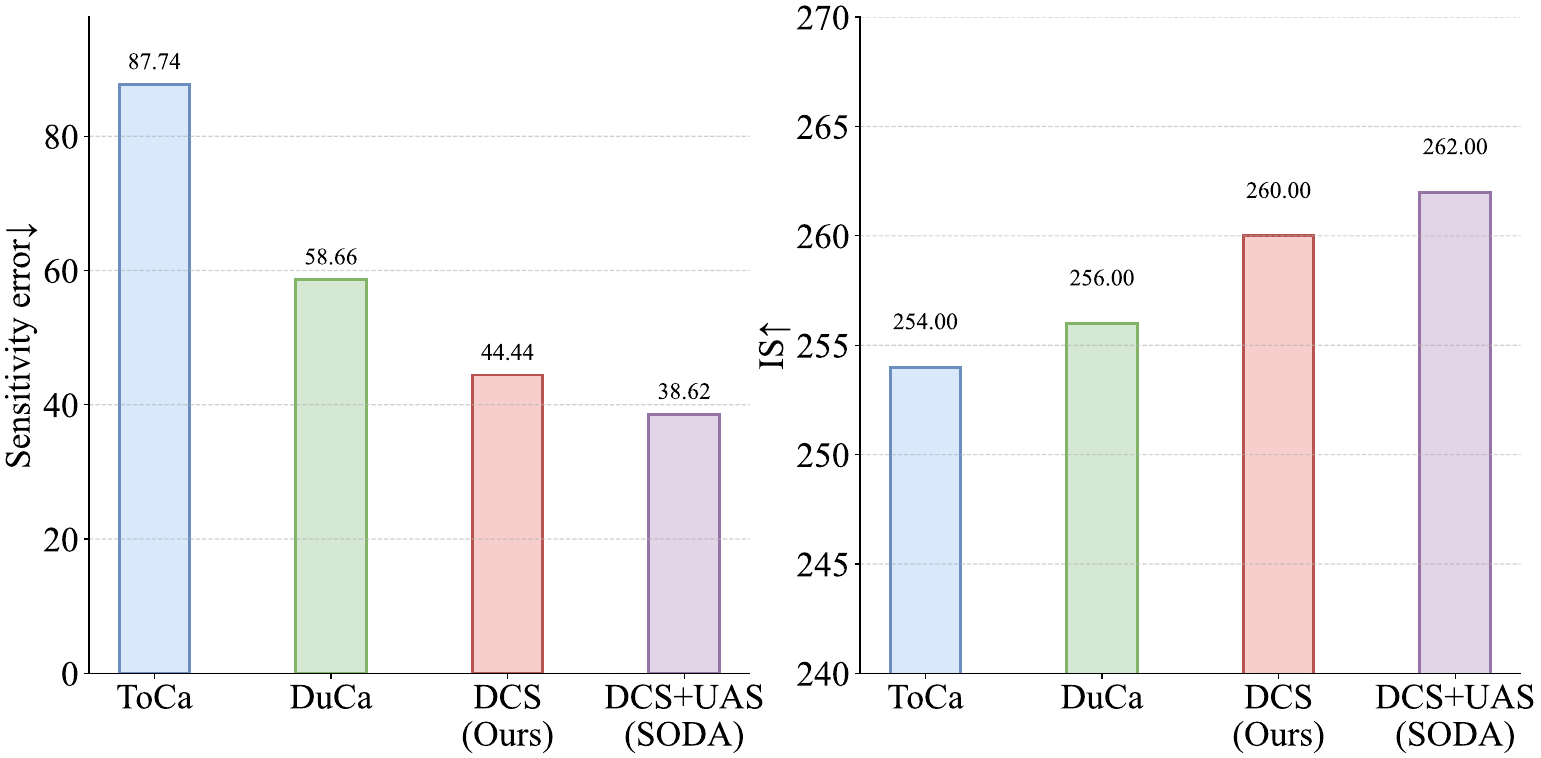} % Reduce the figure size so that it is slightly narrower than the column.
\caption{\textbf{Comparison of cumulative sensitivity error and generation fidelity.} Compared with ToCa and DuCa, our DCS module in SODA effectively reduces sensitivity error under the same acceleration efficiency. When combined with the UAS module, generation fidelity is further improved.}
\label{fig:error_performance}
\end{figure}

\textbf{Adaptive pruning rate.}
Once pruning is triggered, the pruning rate is also adaptively determined by the sensitivity error.
Higher sensitivity errors imply that the corresponding module is more vulnerable to acceleration, thus favoring less pruning (a smaller pruning rate).
Following this intuition, the pruning rate depends on the sensitivity error of the current module.
Still under cache reuse at timestep $t + 1$, our pruning rate is defined as:
\begin{equation}
    \alpha_{t+1,l,m} = \lambda\cdot\mathcal{E}_c(t,l,m,n)+\beta,
\end{equation}
where $\lambda$ is the scaling coefficient, and $\beta$ represents the base pruning rate, which is adaptively adjusted according to the acceleration budget.
It enables the pruning to be simultaneously aware of global budget and local sensitivity.

It should be noted that both adaptive operations rely on offline‑computed sensitivity errors, thus introducing no impact on acceleration efficiency.
While the pruning positions are content‑dependent and computed online, the generation speed reported in Sec.~\ref{sec:exper} already accounts for this overhead.

By integrating the two adaptive modules, SODA builds upon the optimal caching strategy obtained from DCS and further reduces sensitivity error through pruning in UAS, leading to improved generation fidelity, as shown in Fig.~\ref{fig:error_performance}.

\section{Experiments}
\label{sec:exper}
\subsection{Implementation Details}
\mypara{Models and Methods.}
We conduct experiments on DiT-XL/2\cite{peebles2023scalable} (250 DDPM~\cite{ho2020denoising} steps and 50 DDIM steps), PixArt-$\mathrm{\alpha}$\cite{chen2023pixart} (20 DPM++\cite{lu2025dpm} steps), and OpenSora\cite{zheng2024open} (30 reflected flow steps), covering various generation tasks and denoising strategies on A100 and RTX4090 GPUs.

\mypara{Evaluation and Metrics.}
For class-conditional image generation, we use DiT-XL/2 to randomly generate 50k images from ImageNet\cite{deng2009imagenet}, report FID\cite{heusel2017gans}, sFID, IS\cite{salimans2016improved}, Precision, and Recall. 
For text-to-image generation, we report FID-30K from MS-COCO2017\cite{lin2014microsoft} and CLIP score\cite{hessel2021clipscore} generated by PixArt-$\mathrm{\alpha}$.
% FID-30K is calculated between 30,000 generated images and the real images from MS-COCO2017\cite{lin2014microsoft} to assess the generation quality. CLIP score measures the text-image alignment using the 30K text-prompt pairs.
For text-to-video generation, we run VBench\cite{huang2024vbench} with OpenSora.
% Then an evaluation of the 16 aspects proposed by VBench is performed.
We use latency and FLOPs to measure the amount of computation.
More experimental settings can be found in Appendix B.1.

\begin{table*}[!ht]
    \centering
    \scalebox{0.8}{
    \renewcommand\arraystretch{1.0}
    \setlength{\tabcolsep}{12pt}
    \begin{tabular}{cccccccccc}
    \hline
        DiT-XL/2 & Sampling & Latency(s)↓ & FLOPs(T)↓ & Speed↑ & FID↓ & sFID↓ & IS↑ & Precision↑ & Recall↑ \\ \hline
        250 steps & DDPM & 2.508 & 118.68 & 1.00× & 2.23 & 4.57 & 275.65 & 0.82 & 0.58 \\
        \rowcolor{gray!20}\textbf{SODA}($N_s$=125) & DDPM & \textbf{1.778} & \textbf{76.70} & \textbf{1.55×} & \textbf{2.21} & 4.72 & 272.01 & \textbf{0.82} & \textbf{0.58} \\ \hline
        FORA~\cite{selvaraju2024fora} & DDPM & - & 43.36 & 2.74× & 2.80 & 6.13 & / & 0.80 & 0.59 \\
        ToCa~\cite{zou2024accelerating} & DDPM & 1.211 & 43.22 & 2.75× & 2.58 & 5.74 & 256.26 & 0.80 & 0.59 \\
        DuCa~\cite{zou2024accelerating1} & DDPM & 1.170 & 43.42 & 2.73× & 2.59 & 5.68 & 256.36 & 0.80 & 0.59 \\
        \rowcolor{gray!20}\textbf{SODA}($N_s$=72) & DDPM & \textbf{1.070} & 43.42 & 2.73× & \textbf{2.47} & \textbf{5.09} & \textbf{262.30} & \textbf{0.81} & \textbf{0.59} \\ \hline
        % \rowcolor{gray!20}SODA (Ours) & DDPM &  &  &  &  &  &  &  \\ \hline
        50 steps & DDIM & 0.533 & 23.74 & 1.00× & 2.25 & 4.33 & 239.97 & 0.80 & 0.59 \\
        \rowcolor{gray!20}\textbf{SODA}($N_s$=23) & DDIM & \textbf{0.314} & 12.21 & 1.94× & 2.39 & 4.55 & \textbf{240.75} & \textbf{0.81} & 0.58 \\ \hline
        ToCa~\cite{zou2024accelerating} & DDIM & 0.289 & 10.23 & 2.32× & 3.05 & 4.70 & / & 0.79 & 0.57 \\
        DuCa~\cite{zou2024accelerating1} & DDIM & 0.276 & 9.58 & 2.48× & 3.05 & 4.66 & 233.21 & 0.80 & 0.58 \\
        \rowcolor{gray!20}\textbf{SODA}($N_s$=18) & DDIM & 0.263 & 9.55 & 2.49× & \textbf{2.75} & \textbf{4.56} & \textbf{235.65} & \textbf{0.80} & \textbf{0.58} \\
        \rowcolor{gray!20}\textbf{SODA}($N_s$=16) & DDIM & \textbf{0.218} & \textbf{8.43} & \textbf{2.82×} & 2.88 & 4.71 & 234.91  & \textbf{0.80} & \textbf{0.58} \\ \hline
    \end{tabular}}
    \caption{\textbf{Quantitative comparison of DiT-XL/2 on class-conditional image generation.} Gray background indicates our SODA, and bold font highlights the best result. The arrow denotes whether lower or higher values indicate superior performance. $N_s$ represents the caching times and is used to adjust the acceleration budget. Results on more baselines and generation models are provided in Appendix B.2.}
    \label{tab:dit}
\end{table*}

\begin{figure*}[t]
\centering
\includegraphics[width=1.0\textwidth]{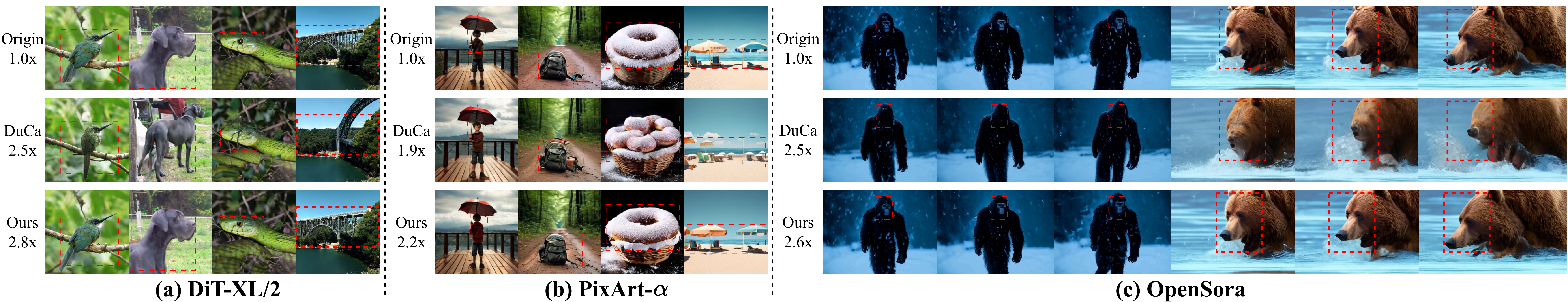} % Reduce the figure size so that it is slightly narrower than the column.
\caption{\textbf{Qualitative comparison of acceleration.} \textbf{(a)} DiT-XL/2 (DDIM). \textbf{(b)} PixArt-$\alpha$. \textbf{(c)} OpenSora. We highlight the areas with red dashed boxes to emphasize the comparison. Our SODA achieves higher generation fidelity under the same or higher acceleration efficiency compared with baselines. More visualizations are provided in Appendix C.
}
\label{fig:visual}
\end{figure*}

\subsection{Main Results}
\mypara{Class-Conditional Image Generation.}
In quantitative results of Tab.~\ref{tab:dit}, SODA improves original performance under low acceleration, achieving the 0.02 FID gain on DDPM and the 0.78 IS increase on DDIM.
At comparable acceleration to the baselines, SODA also enhances generation quality, significantly reducing sFID by 0.65 and increasing IS by over 6 on DDPM.
Moreover, SODA still outperforms the baselines in generation quality at a higher acceleration rate (2.82×).
This is attributed to the effective combination of fine-grained sensitivity modeling and our adaptive strategy.
Sensitivity modeling enables our acceleration to refine decisions at timesteps, layers, and modules, minimizing the error through our adaptive strategy.

Qualitative results also demonstrate its effectiveness. As illustrated in Fig.~\ref{fig:visual}(a), SODA more effectively preserves the generation quality of the original model, maintaining more details under acceleration, such as the snake’s eye in the third column.
The dynamic programming in SODA ensures the optimality of the caching strategy, while adaptive pruning further corrects sensitivity error, thereby improving generation quality under acceleration.

\mypara{Text-to-Image Generation.}
We adopt the same comparative approach in PixArt-$\alpha$, which is shown in Tab.~\ref{tab:pixart}. Under similar acceleration efficiency, SODA achieves a significant 0.72 reduction in COCO-FID compared with DuCa.
As the acceleration rate increases, the improvement in FID diminishes, yet it still achieves optimal performance. Meanwhile, the CLIP score continues to improve, reaching 16.44. Notably, our SODA even significantly outperforms the original model (an improvement of 0.77 in FID, 0.15 in CLIP).

Fig.~\ref{fig:visual}(b) qualitatively demonstrates that SODA achieves better content consistency with the original model and maintains image details.
For instance, the preservation of the boy's background in the first column and the number of donuts in the third column demonstrate that SODA better recovers the original model's fidelity, without generating other content due to stochastic deviations.

\begin{figure}[t]
\centering
\includegraphics[width=0.4\textwidth]{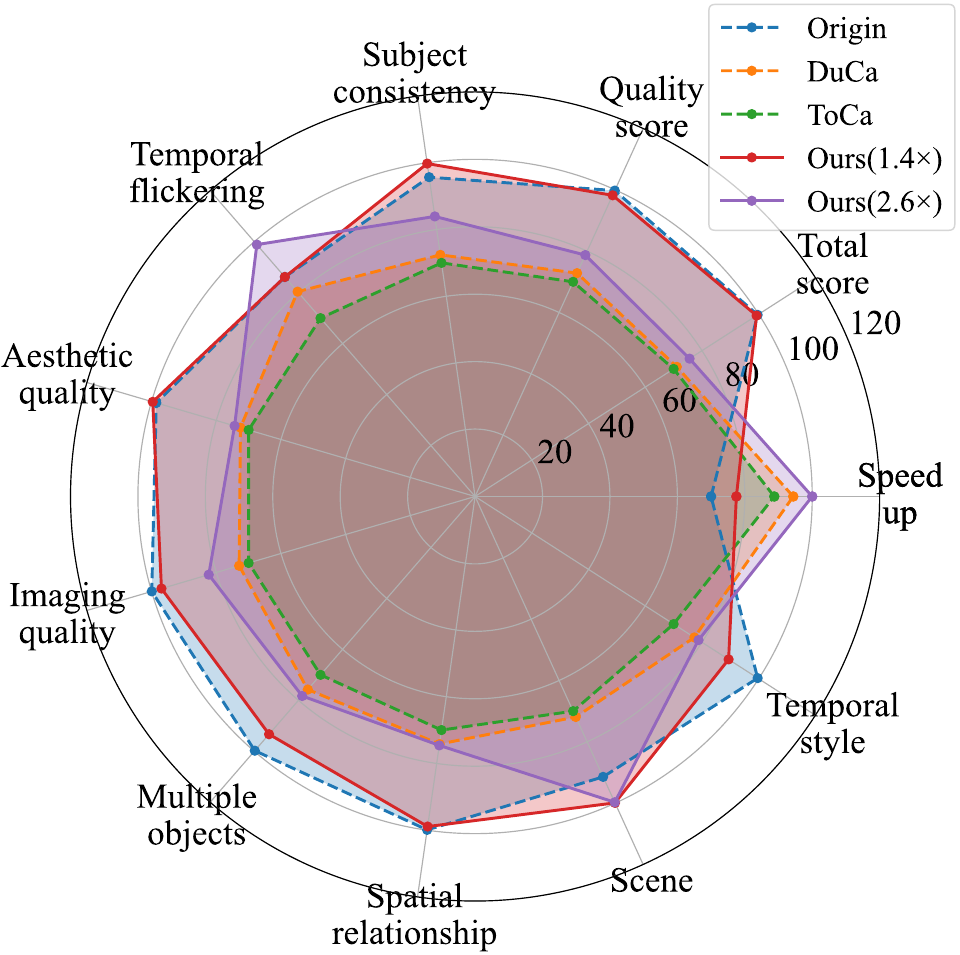} % Reduce the figure size so that it is slightly narrower than the column.
\caption{\textbf{VBench detailed metric comparison.} We normalize metrics by setting the highest value to 100\%, and visualize 9 selected metrics along with acceleration efficiency.}
\label{fig:vbench}
\end{figure}

\mypara{Text-to-Video Generation.}
In addition to image generation, we also evaluate SODA on video generation.
In Tab.~\ref{tab:opensora}, SODA maintains generation quality at 1.42× acceleration without observable degradation.
Under comparable FLOPs with baselines, SODA further improves the average performance by 0.1 on the VBench, and achieves optimal performance on the fine-grained metrics presented in Fig.~\ref{fig:vbench}.

In contrast to image generation, video generation necessitates the modeling of temporal dependencies across frames.
The results demonstrate that our sensitivity modeling is also capable of assessing the acceleration sensitivity when generating videos.
By further incorporating our adaptive acceleration, SODA facilitates a seamless transition of acceleration automatically from image to video.

\begin{table}[t]
    \centering
    \scalebox{0.8}{
    \begin{tabular}{cccccc}
    \hline
        \multirow{2}{*}{PixArt-$\alpha$} & \multirow{2}{*}{\makecell[c]{Latency\\(s)↓}} & \multirow{2}{*}{\makecell[c]{FLOPs\\(T)↓}} & \multirow{2}{*}{Spe.↑} & \multicolumn{2}{c}{\makecell[c]{MS-COCO2017\\w/ difference(‰)}} \\ \cline{5-6}
        & & & & FID-30k↓ & CLIP↑ \\ \hline
        20 steps & 0.857 & 11.18 & 1.00× & 28.10 & 16.29 \\
        10 steps & 0.491 & 5.54 & 2.00× & 37.57 & 15.83 \\
        FORA(ArXiv24) & - & 5.66 & 1.98× & 29.67\textsubscript{55.9↓} & 16.40\textsubscript{6.7↑} \\
        ToCa(ICLR25) & 0.492 & 6.05 & 1.85× & 28.38\textsubscript{9.9↓} & 16.43\textsubscript{8.6↑} \\
        DuCa(ArXiv24) & 0.479 & 5.99 & 1.87× & 28.05\textsubscript{1.8↑} & 16.42\textsubscript{8.0↑} \\
        \rowcolor{gray!20}\textbf{SODA}($N_s$=8) & 0.472 & 5.95 & 1.88× & \textbf{27.33}\textsubscript{\textcolor{red}{27.4↑}} & 16.42\textsubscript{8.0↑} \\
        \rowcolor{gray!20}\textbf{SODA}($N_s$=7) & \textbf{0.428} & \textbf{5.07} & \textbf{2.21×} & 27.72\textsubscript{13.5↑} & \textbf{16.44}\textsubscript{\textcolor{red}{9.2↑}} \\ \hline
    \end{tabular}}
    \caption{\textbf{Quantitative comparison of PixArt-$\boldsymbol{\alpha}$ on text-to-image generation.} Subscripts indicate the relative difference compared to the original results. Values in red denote the best performance.}
    \label{tab:pixart}
\end{table}

\begin{table}[!ht]
    \centering
    \scalebox{0.86}{
    \begin{tabular}{ccccc}
    \hline
        OpenSora & \makecell[c]{Latency\\(s)↓} & \makecell[c]{FLOPs\\(T)↓} & Spe.↑ & \makecell[c]{VBench(\%↑)\\w/ degrad(‰)} \\ \hline
        30 steps & 102.287 & 3283.20 & 1.00× & 79.13 \\ 
        \rowcolor{gray!20}\textbf{SODA}($N_s$=21) & \textbf{86.684} & \textbf{2319.94} & \textbf{1.42×} & \textbf{79.13}\textsubscript{\textcolor{red}{0.00↓}} \\ \hline
        PAB(ICLR25) & - & 2558.25 & 1.28× & 76.95\textsubscript{27.55↓} \\ 
        FORA(ArXiv24) & - & 1751.32 & 1.87× & 76.91\textsubscript{28.06↓} \\ 
        $\Delta$-DiT(ArXiv24) & - & 3166.47 & 1.04× & 78.21\textsubscript{11.62↓} \\
        ToCa(ICLR25)& 54.698 & 1394.03 & 2.36× & 78.34\textsubscript{9.98↓} \\ 
        DuCa(ArXiv24)& 50.637 & 1315.62 & 2.50× & 78.39\textsubscript{9.35↓} \\
        \rowcolor{gray!20}\textbf{SODA}($N_s$=12) & 49.896 & 1314.42 & 2.50× & \textbf{78.49}\textsubscript{\textcolor{red}{8.08↓}} \\
        \rowcolor{gray!20}\textbf{SODA}($N_s$=11) & \textbf{48.022} & \textbf{1277.59} & \textbf{2.57×} & 78.41\textsubscript{9.09↓} \\ \hline
    \end{tabular}}
    \caption{\textbf{Quantitative comparison of OpenSora on text-to-video generation.} Subscripts indicate the relative degradation compared to the original results.}
    \label{tab:opensora}
\end{table}

\begin{table}[t]
    \centering
    \scalebox{0.87}{
    \begin{tabular}{cccccccc}
    \hline
        Modules & FID↓ & sFID↓ & IS↑ & Precision↑ & Recall↑ \\ \hline
        Vanilla & 3.83 & 5.24 & 213.12 & 0.77 & 0.57 \\
        OFS w/ DCS & 2.78 & 4.63 & 234.78 & 0.80 & 0.58\\
        OFS w/ UAS & 2.89 & 4.75 & 235.01 & 0.79 & 0.58\\
        \rowcolor{gray!20} \textbf{SODA}(Ours) & \textbf{2.75} & \textbf{4.56} & \textbf{235.65} & \textbf{0.80} & \textbf{0.58}\\ \hline
        L1 & 2.92 & 4.91 & 233.62 & 0.79 & 0.58 \\
        L2 & 2.88 & 4.88 & 233.91 & 0.79 & 0.58\\
        \rowcolor{gray!20} \textbf{COS}(Ours) & \textbf{2.75} & \textbf{4.56} & \textbf{235.65} & \textbf{0.80} & \textbf{0.58}\\ \hline
    \end{tabular}}
    \caption{\textbf{Ablation study for different modules under 2.5× acceleration on DiT-XL/2 with DDIM.} ``Vanilla'' is the naive caching with fixed cache intervals. ``L1'', ``L2'', and ``COS'' denote the metrics used to define sensitivity error in OFS module.}
    \label{tab:ablation}
\end{table}

In the qualitative results, SODA also preserves more details consistent with the original outputs, such as the face of the bigfoot in Fig.~\ref{fig:visual}(c). It further demonstrates the effectiveness of combining caching based on dynamic programming with adaptive pruning decision in SODA.

\subsection{Ablation Studies}

We analyze the effectiveness of Offline Fine-grained Sensitivity Modeling (OFS), Unified Adaptive Strategy Formulation (UAS), and Dynamic Caching Scheduling Optimization (DCS) on DiT-XL/2 with DDIM. More ablation studies are provided in Appendix B.3.

\subsubsection{Module effectiveness}
To validate the effectiveness of each module, we report the performance with individual modules in Tab.~\ref{tab:ablation}. Since the OFS serves as the prior for the subsequent components, it is only meaningful to evaluate it with other modules.

The DCS treats sensitivity error as dynamic cost and leverages dynamic programming to derive the optimal caching that minimizes cumulative error.
As a result, DCS enhances generation quality during acceleration, with the FID dropping by 1.05 and the IS increasing by 21.66 significantly.
UAS mitigates the error caused by long cache intervals through adaptive pruning decision.
By synchronizing the pruning rates with the error distribution, UAS retains computations that are more sensitive to caching, correcting caching errors. It allows the DiT to achieve acceleration while simultaneously preserving the quality, achieving the FID dropping by 0.97 and the IS increasing by 21.89.

The combination of DCS and UAS reduces the minimum cumulative error, thereby boosting model performance while achieving efficient acceleration.
In addition, Tab.~\ref{tab:ablation} further compares different modeling approaches for sensitivity error.
COS in OFS achieves the best fidelity.

% \begin{table}[t]
%     \centering
%     \scalebox{0.87}{
%     \setlength{\tabcolsep}{10pt}
%     \begin{tabular}{cccccc}
%     \hline
%     \# of samples & 10 & 50 & 100 & 500 & 1000 \\ \hline
%     FID↓ & & & 2.75 & & 2.71\\
%     IS↑ & & & 235.65 & & 236.04 \\ \hline
%     \end{tabular}}
%     \caption{\textbf{Offline cost of different models.} \# denotes the number. The cost of OFS includes the cost of random content generation (the cost of only generation).}
%     \label{tab:num}
% \end{table}

\subsubsection{Offline analysis}
\label{sec:offline}
As the core of SODA is its offline sensitivity modeling in OFS module, we conduct additional analysis focusing on several key aspects: (1) the cost of OFS, (2) the consistency of sensitivity distributions between offline modeling and actual inference, and (3) the effect of the number of random samples used in offline modeling on performance.

\textbf{Offline Cost.}
To balance the efficiency of offline modeling and actual performance, we use 100 random samples for averaging in image models and 10 in video models.
The offline overhead is summarized in Tab.~\ref{tab:cost}, all time within one hour.
Since each model requires only a single offline modeling process, the offline cost of SODA is extremely low.
In addition, since extra caches need to be stored during offline modeling, the memory consumption is slightly higher than that of normal inference.
However, during online inference, only the precomputed sensitivity error results are loaded, and the memory overhead is therefore negligible.

\begin{figure}[t]
\centering
\includegraphics[width=0.48\textwidth]{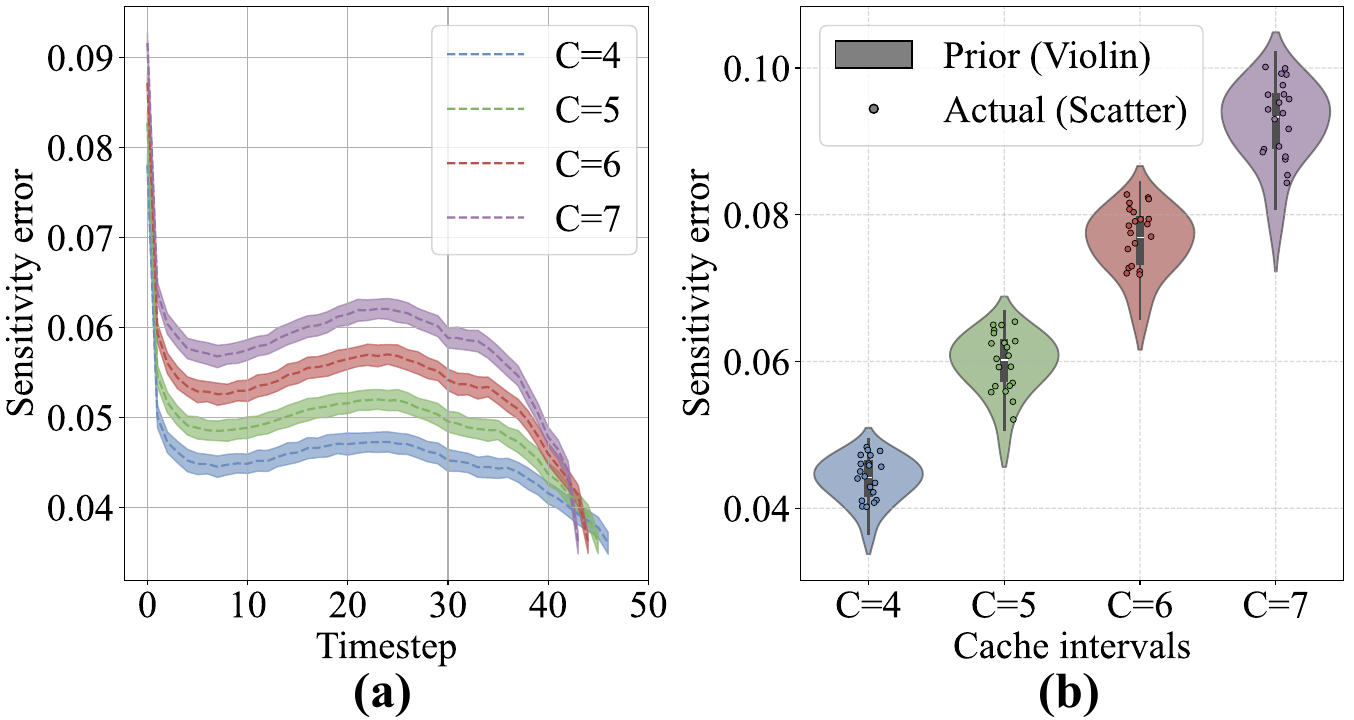} % Reduce the figure size so that it is slightly narrower than the column.
\caption{\textbf{Analysis of sensitivity error distribution.}
\textbf{(a)} Mean–variance of offline sensitivity errors under different cache intervals. The variance remains stable and low across multiple random generations, indicating that sensitivity is primarily model‑related and marginally affected by content changes.
\textbf{(b)} Consistency of sensitivity error distributions between offline modeling and actual inference under different cache intervals. It reveals that the sensitivity error distributions obtained from offline modeling are consistent with those during actual inference.}
\label{fig:distribution}
\end{figure}

\begin{table}[t]
    \centering
    \scalebox{0.78}{
    \begin{tabular}{ccccccc}
    \hline
    Phase & Cost & DiT-XL/2 & PixArt-$\alpha$ & OpenSora \\ \hline
    & \# of samples & 100 & 100 & 10 \\ \hline
    \multirow{4}{*}{\makecell[c]{Offline\\modeling}} & Time of generation & 69.71s & 90.13s & 24.23m \\
    & Time of OFS & 91.23s & 98.71s & 31.03m \\
    & Memory of generation & 4.09GB & 22.31GB & 52.40GB \\
    & Memory of OFS & 4.65GB & 22.65GB & 52.40GB \\ \hline
    Online use & Memory of $\mathcal{E}$ & 0.16MB & 0.13MB & 0.27MB \\ \hline
    \end{tabular}}
    \caption{\textbf{Offline cost of different models.} \# denotes the number. The cost of OFS includes the cost of random content generation (the cost of generation). This modeling process needs to be performed only once per generation model and can be permanently reused, independent of variations in CFG, seed, or prompt.
At runtime, the sensitivity errors $\mathcal{E}$ are simply loaded into memory, introducing no time overhead and negligible memory cost.}
    \label{tab:cost}
\end{table}

\textbf{Number of offline random samples.}
In addition, the representativeness of random samples during offline modeling affects the generation quality under acceleration, which is closely related to the number of random samples used.
Taking images as an example, we evaluated generation fidelity of DiT-XL/2 based on sensitivity errors across 100 and 1000 random generations, respectively.
The corresponding FID↓ is 2.75 and 2.71, and the IS↑ is 235.65 and 235.74, showing almost no difference.
The current setting of 100 balances modeling efficiency and generation quality. These results demonstrate that the current number of random generations is sufficiently representative and further suggest that the observed sensitivity is inherent to the model rather than related to the generated content.

\textbf{Distribution consistency.}
Finally, we are concerned with whether the sensitivity error distribution obtained offline is consistent with that observed during actual inference.
To investigate this, we use violin plots to depict the offline sensitivity error distribution and scatter points to represent the inference-time errors, aiming to examine whether the latter falls within the offline error distribution.
As shown in Fig.~\ref{fig:distribution}, the results demonstrate a high degree of consistency between the two distributions, confirming the effectiveness of the offline method in modeling sensitivity while avoiding online computational overhead.

\subsubsection{Hyper-parameters}
\begin{figure}[t]
\centering
\includegraphics[width=0.46\textwidth]{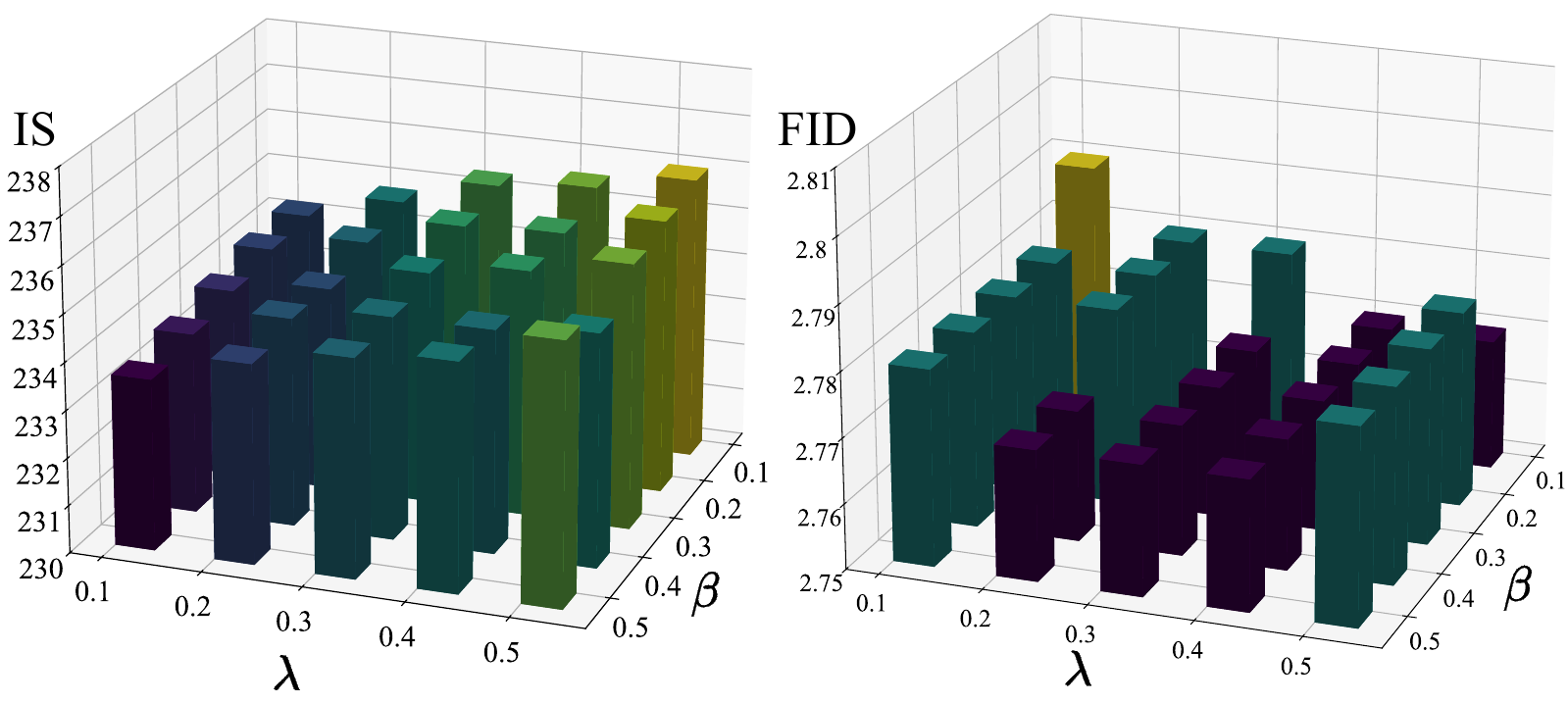} % Reduce the figure size so that it is slightly narrower than the column.
\caption{\textbf{The impact of parameters $\lambda$ and $\beta$ on DiT-XL/2 with DDIM.} $\Delta$FID$\leq$0.02, and $\Delta$IS$\leq$2.2.} 
\label{fig:params}
\end{figure}

In addition, UAS introduces two hyper-parameters, $\beta$ and $\lambda$, which govern the base ratio and the scaling of relative sensitivity differences, respectively.
They introduce minor fluctuations in FLOPs and quality performance. To investigate their sensitivity and influence, we present a visualization of the parameter space in Fig.~\ref{fig:params}.
The results exhibit minimal fluctuation across different parameter settings, suggesting that the model is not highly sensitive to parameter changes.

% \subsubsection{Flexible Budgets.}
% SODA allows for controlling acceleration efficiency by adjusting the number of cache interval $N_s$, as illustrated in Fig.~\ref{fig:head2}. It achieves the balance between acceleration and generation quality, and can even boost the model's original performance under low acceleration ratio.

% In addition, EAAP introduces two hyper-parameters, $\beta$ and $\lambda$, which govern the base ratio and the scaling of relative error differences, respectively.
% They introduce minor fluctuations in flops and model performance. To investigate their sensitivity and influence, we present a visualization of the parameter space in Fig.~\ref{fig:params}.
% The results exhibits minimal fluctuation across different parameter settings, suggesting that the model is not highly sensitive to parameter changes.

\section{Concluding Remarks}
\mypara{Summary.}
We present SODA, a Sensitivity‑Oriented Dynamic Acceleration framework for DiT that unifies caching and pruning through fine‑grained sensitivity modeling. SODA establishes the offline sensitivity error modeling across timesteps, layers, and modules, enabling dynamic programming–based cache optimization and adaptive pruning of insensitive tokens. It effectively balances acceleration efficiency and generation fidelity. Extensive experiments demonstrate that it achieves controllable acceleration with higher fidelity and high cross-model generalization.

\mypara{Limitations.}
SODA achieves training-free adaptive acceleration for DiT. However, its effect lags behind training-based methods. Moreover, while SODA is compatible with various models, further exploration is needed to investigate its integration with techniques like distillation.

\subsection*{Acknowledgments}

This work was supported by National Natural Science Foundation of China (grant No. 62350710797), by Guangdong Basic and Applied Basic Research Foundation (grant No. 2023B1515120065, 2025A1515011546), and by the Shenzhen Science and Technology Program
(JCYJ20240813105901003, KJZD20240903102901003, ZDCY20250901113000001).

{
    \small
    \bibliographystyle{ieeenat_fullname}
    \bibliography{main}
}

% \clearpage
\setcounter{page}{1}
\maketitlesupplementary
\appendix
% 相关工作
% 离线误差分析
% 动态规划缓存
% 实验部分：实验细节设置，更多实验结果，vbench结果细节，更多可视化

\section*{Overview}
This is the appendix for our paper titled \textit{SODA: Sensitivity-Oriented Dynamic Acceleration for Diffusion Transformer}. This appendix supplements the main paper with the following content:
\begin{itemize}
  \renewcommand{\labelitemi}{$\bullet$}
  \item \ref{soda} \textbf{More Details of our SODA}
  \begin{itemize}
    \item[--] \ref{ofs} Details of Offline Fine-grained Sensitivity Modeling (OFS)
    \item[--] \ref{dcs} Details of Dynamic Caching Scheduling Optimization (DCS)
    \item[--] \ref{uas} Details of Unified Adaptive Strategy Formulation (UAS)
    \item[--] \ref{overall} Overall Analysis
  \end{itemize}
  \item \ref{exper} \textbf{More Experiments}
  \begin{itemize}
    \item[--] \ref{detail} More Implementation Details
    \item[--] \ref{results} More Experiment Results
    \item[--] \ref{ablation} More Ablation Study
  \end{itemize}
  \item \ref{visualization} \textbf{More Visualization}
  \begin{itemize}
    \item[--] \ref{visual_dit} DiT-XL/2
    \item[--] \ref{visual_pixart} PixArt-$\alpha$
    \item[--] \ref{visual_opensora} OpenSora
  \end{itemize}
  \item \ref{related} \textbf{Related Work}
  \begin{itemize}
    \item[--] \ref{dit} Diffusion Transformer
    \item[--] \ref{acceleration} Training-free DiT Acceleration
    \item[--] \ref{others} Other Acceleration
  \end{itemize}
\end{itemize}

\section{More Details of our SODA}
\label{soda}
To adapt to the complex internal sensitivity to acceleration, we propose SODA, a Sensitivity-Oriented Dynamic Acceleration method.
As illustrated in Fig.~\ref{fig:model}, SODA first conducts
\textbf{(1) Offline Fine-grained Sensitivity Modeling (OFS)}: Defining error to measure the fine-grained sensitivity of timesteps, layers and modules before inference.
Then, SODA adopts \textbf{(2) Dynamic Caching Scheduling Optimization (DCS)}: Employing dynamic programming to identify the optimal combination of cache intervals that yields the minimal sensitivity impact.
Finally, when pruning and cache reuse, SODA proposes
\textbf{(3) Unified Adaptive Strategy Formulation (UAS)}: Achieving adaptive scheduling for pruning timing and rate guided by sensitivity errors.

Due to space limitations in the main text, we provide additional analysis and implementation details of this module here.

\subsection{Details of Offline Fine-grained Sensitivity Modeling (OFS)}
\label{ofs}

Although the integration of caching and pruning balances the high acceleration of cache with the flexibility of pruning, existing fixed or heuristic methods struggle to accurately perceive the sensitivity introduced by such acceleration operations. As a result, they inevitably skip important computations, leading to degraded generation quality.

\subsubsection{Sensitivity error and generation fidelity.}
We emphasize the importance of the sensitivity error because it is the fundamental cause of the degradation in generation quality.

Once an acceleration strategy is determined, the diffusion inference results at each timestep and layer exhibit slight deviations due to acceleration operations such as cache reuse or pruning, thereby introducing computational errors.
As the inference process unfolds, minor errors introduced by acceleration operations accumulate rapidly, often in an exponential manner. This accumulation results in substantial deviations from the ground-truth outputs, severely compromising generation quality.
The essence of current acceleration optimization strategies lies in minimizing the acceleration-induced errors as much as possible, so that the accelerated diffusion process can closely approximate the original diffusion outputs.

However, the introduction of such errors is influenced by the model’s intrinsic sensitivity to acceleration, which is highly complex and variable. The degree of sensitivity differs across models, timesteps, layers, and modules, making it difficult to capture through prior knowledge or fixed heuristic rules. As a result, existing static or heuristic acceleration strategies fail to fully account for these sensitivity‑induced errors and inevitably lead to performance degradation.

To address this challenge, we propose a sensitivity quantification approach that models the fine‑grained, dynamic distribution of sensitivity errors within different modules, enabling a more precise characterization of the model’s response to acceleration.

\begin{figure*}[t]
\centering
\includegraphics[width=0.65\textwidth]{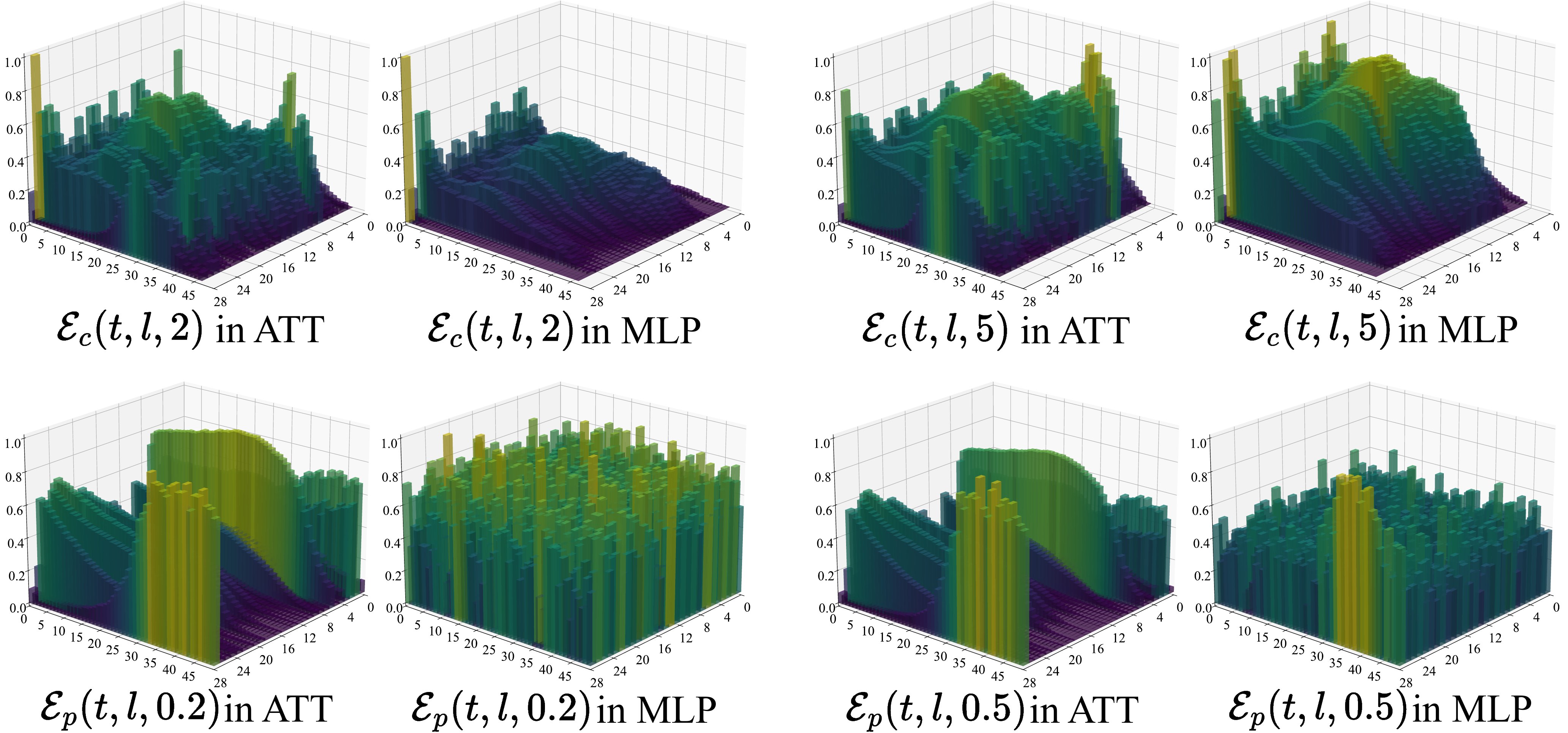} % Reduce the figure size so that it is slightly narrower than the column.
\caption{\textbf{Sensitivity error visualization of DiT-XL/2.} We visualize sensitivity errors of each module (attention is abbreviated as ATT) at every timestep and layer during the denoising process, under different cache intervals and pruning rates.}
\label{fig:dit_error}
\end{figure*}

\begin{figure*}[t]
\centering
\includegraphics[width=0.9\textwidth]{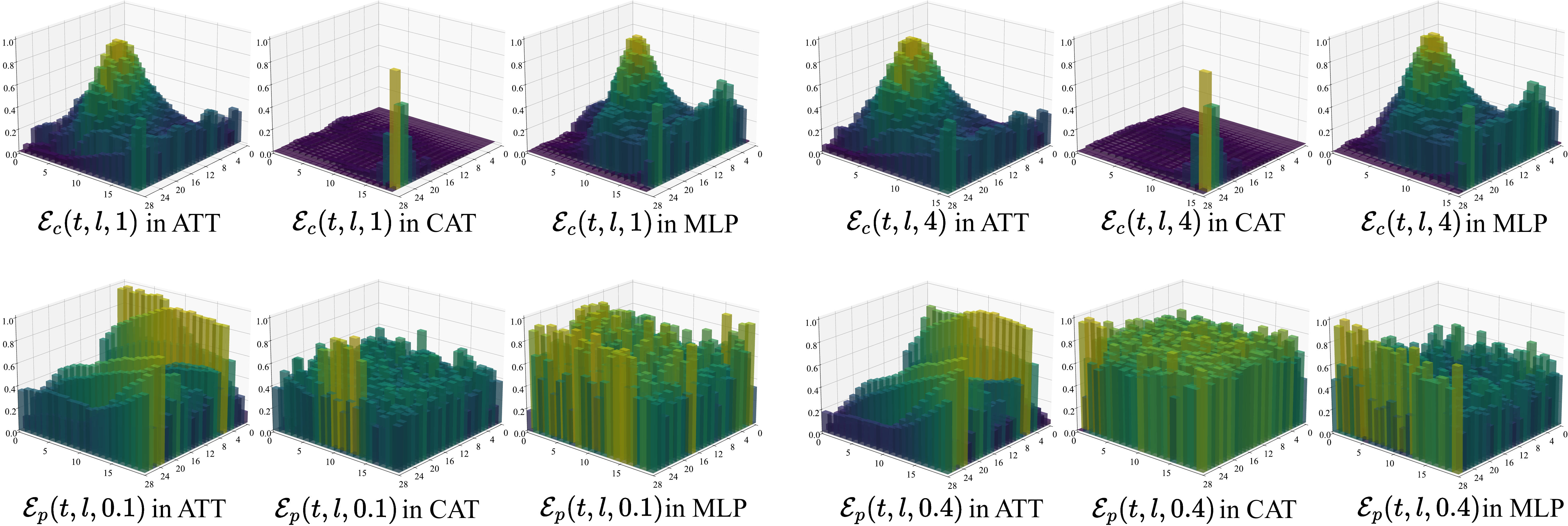} % Reduce the figure size so that it is slightly narrower than the column.
\caption{\textbf{Sensitivity error visualization of PixArt-$\alpha$.} We visualize the sensitivity error of each module (attention is abbreviated as ATT, cross-attention as CAT) at every timestep and layer during the denoising process, under different cache intervals and pruning rates.}
\label{fig:pixart_error}
\end{figure*}

\subsubsection{Purpose of sensitivity error modeling.}
The essential goal of our sensitivity error analysis is to:
\begin{itemize}
    \item Through error analysis, we reveal that sensitivity varies significantly across different models, timesteps, layers, and modules.
    \item It highlights the limitations of current heuristic-based approaches and underscores the urgent need for adaptive methods that leverage the sensitivity error distribution to better balance acceleration and output quality.
    \item Based on the estimated sensitivity errors, we derive the optimal caching strategy through dynamic programming prior to inference by minimizing the cumulative sensitivity error. This not only reduces sensitivity errors but also avoids any overhead during the actual inference.
    \item By comparing the estimated pruning sensitivity error with the caching sensitivity error, we adaptively determine when to apply pruning operations. It allows us to retain a subset of tokens to mitigate sensitivity errors, thus achieving acceleration while simultaneously correcting for cache-induced quality degradation.
\end{itemize}

\subsubsection{Details of sensitivity error.}
As described in the main paper, we model the sensitivity errors caused by cache operations ($\mathcal{E}_c$) and pruning ($\mathcal{E}_p$) based on the Cosine distance between the accelerated outputs and the ground-truth outputs.

We compute both $\mathcal{E}_c$ and $\mathcal{E}_p$ separately across different models and modules. For $\mathcal{E}_c$, we evaluate the error under cache intervals ranging from 1 to 9. In contrast, since the pruning rate is a floating-point value rather than a discrete integer like the cache interval, we analyze the general trend of $\mathcal{E}_p$ by varying the pruning rate from 0.1 to 0.9 with a step size of 0.1.

During the detailed analysis, $\mathcal{E}_c$ is influenced by multiple variables: the models (DiT-XL/2, PixArt-$\alpha$, OpenSora), modules (ATT, CAT, MLP), timesteps ($1-T$), layers ($1-L$), and cache intervals ($1-9$).
$\mathcal{E}_p$ follows the same pattern.
When analyzing a specific dimension, we average all other factors to observe the overall trend. For example, when comparing the error distribution across different models, we average the layers, modules, and operations, retaining only the model and timestep dimensions for comparison.

\begin{figure}[t]
\centering
\includegraphics[width=0.35\textwidth]{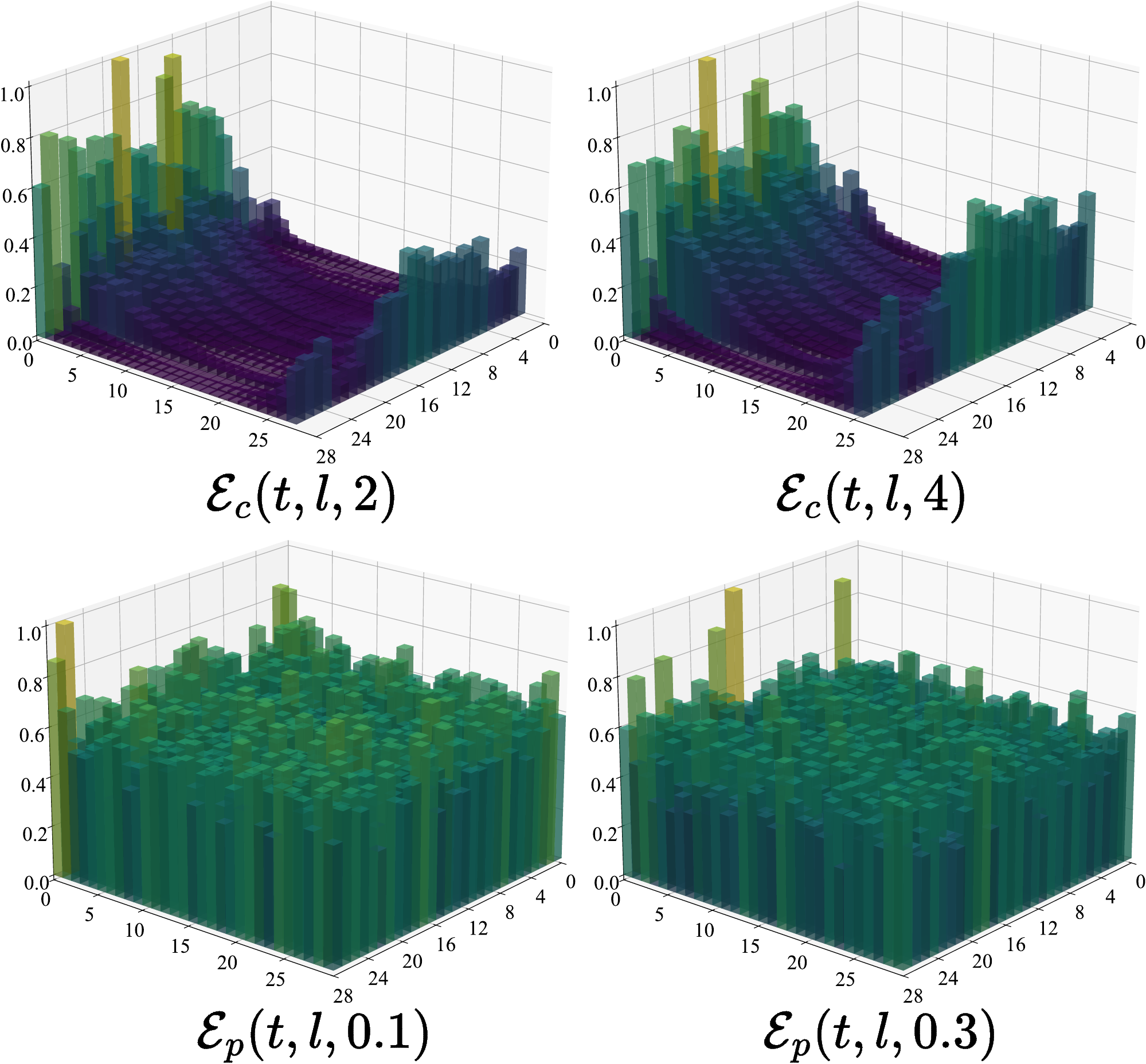} % Reduce the figure size so that it is slightly narrower than the column.
\caption{\textbf{Sensitivity error visualization of MLP in OpenSora.} We visualize sensitivity errors of each module at every timestep and layer during the denoising process, under different cache intervals and pruning rates.}
\label{fig:opensora_error}
\end{figure}

\subsubsection{More sensitivity analysis.}
We perform sensitivity error estimation analyses on DiT-XL/2, PixArt-$\alpha$, and OpenSora to cover a wide range of tasks and model architectures.

In DiT, we visualize two different acceleration ratios under various acceleration operations, and compare the results across different modules in Fig.~\ref{fig:dit_error}.
It can be observed that even under the same acceleration operation, different modules exhibit significantly different sensitivity error distributions. For example, attention modules are more sensitive to acceleration, while modules that do not emphasize contextual information like MLP tend to present error patterns that are nearly random under pruning strategy.

Furthermore, the discrepancy across different acceleration operations is significant. As a result, for our approach that combines caching and pruning, analyzing the sensitivity error characteristics of each acceleration strategy is critically important.

Similar analyses are performed on PixArt and OpenSora, where comparable discrepancies can be observed, as illustrated in Fig.~\ref{fig:pixart_error} and Fig.~\ref{fig:opensora_error}.
By comparing the analysis results across different models, we observe that while there are certain commonalities along specific dimensions (PixArt-$\alpha$ tends to exhibit higher sensitivity errors from caching-based acceleration in the middle stages of denoising, whereas OpenSora shows more concentrated errors at the beginning and end), these patterns are difficult to generalize at each module level. As a result, current heuristic approaches inevitably introduce relatively high sensitivity errors, which in turn lead to noticeable degradation in output quality.

Therefore, based on the sensitivity error estimation, our analysis leads to the following conclusions:
\begin{itemize}
    \item Diffusion models exhibit highly diverse sensitivity error behaviors under different acceleration operations, with significant variations across models, timesteps, layers, and modules. Such complex sensitivity error distributions are difficult to accurately capture through manually crafted heuristics.
    \item The observed heterogeneity persists across diverse image and video generation tasks, regardless of content differences. This consistency supports its use as a prior for guiding inference-time decisions.
\end{itemize}

\begin{figure}[t]
\centering
\includegraphics[width=0.4\textwidth]{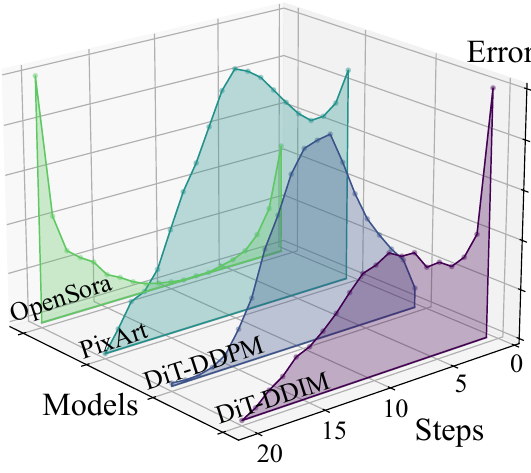} % Reduce the figure size so that it is slightly narrower than the column.
\caption{\textbf{Sensitivity error visualization of different models.} There exist distinct phases where the sensitivity error reduction is particularly pronounced in every model.}
\label{fig:model_error}
\end{figure}

\subsection{Details of Dynamic Caching Scheduling Optimization (DCS)}
\label{dcs}

We propose a caching strategy formulated as a dynamic programming problem, where sensitivity errors are treated as step-wise costs. This allows us to adaptively determine the optimal caching intervals and anchor timestep sets that minimize the overall accumulated error.

The main idea and procedure of the algorithm have already been presented in the main paper.
Here, we provide a supplementary explanation of the proposed algorithm, demonstrating how the dynamic programming procedure can be flexibly adapted to accommodate specific practical scenarios.

During the adaptive caching process, we observe that the DiT architecture typically exhibits a special stage in which the introduced error is abnormally reduced.
As shown in Fig.~\ref{fig:model_error}, this phenomenon is consistently observed across all models in our experiments. This suggests that the DiT architecture possesses an inherent error-correction capability and demonstrates a degree of robustness to introduced sensitivity errors.

Since this special phase varies significantly across different models, we adopt an adaptive approach to incorporate this extension, in order to ensure the generality of our method.

We first compute the gradient $\nabla \mathcal{E}_{dp}$ of the step-wise cost $\mathcal{E}_{dp}$. After smoothing, this special stage can be identified by simply determining the sign of the gradient $\nabla \mathcal{E}_{dp}$. Within this identified stage, the candidate cache interval set $\mathcal{N}$ is restricted as $\mathcal{N}'$ by favoring smaller cache intervals, thus preserving the inherent robustness of DiT to perturbations.

With this approach, we are able to explicitly retain more timesteps that fall within this special phase during the dynamic programming process, thereby preserving the inherent noise-suppression capability of the DiT architecture.

The extended algorithm, as shown in Alg.~\ref{alg:dp2} (we highlight the extended parts in blue font), primarily narrows the search space during the optimal substructure search by introducing targeted constraints on specific timesteps.
Specifically, suppose that the original candidate set $\mathcal{N}=\{n|n\in \mathbb{Z},2\leq n \leq 6\}$. We can define a reduced set $\mathcal{N}'=\{n|n\in \mathbb{Z},2\leq n \leq 3\}$, which contains only half of the elements of the original one. This allows the dynamic programming process to adopt a more conservative caching interval during the special phase.

By extending this idea, we demonstrate the flexible application of dynamic programming in caching strategies, contributing to a deeper integration of this mathematical framework with DiT-based acceleration. We plan to further investigate this integration in future work.

\begin{algorithm}[t]
\renewcommand{\algorithmicrequire}{\textbf{Input:}}
\renewcommand{\algorithmicensure}{\textbf{Output:}}
\caption{Dynamic Caching Scheduling Optimization}
\label{alg:dp2}
\begin{algorithmic}[1]
\REQUIRE Total steps $T$, caching times $N_s$, cache interval candidate set $\mathcal{N}$, \textcolor{blue}{constrained cache interval candidate set $\mathcal{N}'$}, sensitivity error here $\mathcal{E}_{dp}(t,n)$
\ENSURE timestep set $\mathcal{A}$ and cache interval set $\mathcal{I}$
\STATE Initialize $dp[t][i]\gets \infty$, $dp[T][1]\gets 0$, $prev[t][i]\gets \text{None}$, $\forall t\in[T,1],\forall i\in[1,N_s]$ \\
\FOR{$i = 1$ to $N_s$}
    \FOR{$t = T$ down to $1$}
        \IF{\textcolor{blue}{$\Delta \mathcal{E}_{dp}(t,n)<0$}}
            \STATE \textcolor{blue}{$\mathcal{N}^* \gets \mathcal{N}'$}
            \ELSE
            \STATE \textcolor{blue}{$\mathcal{N}^* \gets \mathcal{N}$}
        \ENDIF
        % \IF{$dp[t][i] < \infty$}
            \FORALL{$n \in \textcolor{blue}{\mathcal{N}^*}$}
                % \STATE $t_{\text{next}} \gets t - n$
                \IF{$t > 0$}
                    % \STATE $cost \gets \mathcal{E}_a(t, n)$
                    \IF{$dp[t][i+1] > dp[t+n][i] + \mathcal{E}_{dp}(t,n)$}
                        \STATE $dp[t][i+1] \gets dp[t+n][i] + \mathcal{E}_{dp}(t,n)$
                        \STATE $prev[t][i+1] \gets (t, n)$
                    \ENDIF
                \ENDIF
            \ENDFOR
        % \ENDIF
    \ENDFOR
\ENDFOR
\STATE Backtracking: $\mathcal{A}\gets \{\}$, $\mathcal{I}\gets \{\}$, $t \gets 1, i\gets N_s$
% \STATE Initialize: cache interval set $\mathcal{I}\gets \{\}$,
% \STATE Backtracking start: $t \gets 1, \quad k \gets N_s$
\WHILE{$i > 0$}
    \STATE $(t_{\text{next}}, n) \gets \text{prev}[t][i]$
    \STATE $\mathcal{A} \gets \mathcal{A} \cup \{t\}$, $\mathcal{I} \gets \mathcal{I} \cup \{n\}$
    \STATE $t \gets t_{\text{next}}$, $i \gets i - 1$
\ENDWHILE
% \STATE Reverse $anchors$ and $intervals$
\RETURN $\mathcal{A}$, $\mathcal{I}$
\end{algorithmic}
\end{algorithm}

\subsection{Details of Unified Adaptive Strategy Formulation (UAS)}
\label{uas}
In the main paper, we have already explained whether pruning is applied at each timestep and the corresponding pruning rate:
\begin{equation}
x'=\left\{
             \begin{array}{lr}
             \mathcal{C}_{l,m}(\Omega),\ if\ \delta_{t+1,l,m}\geq\mathcal{E}_c(t,l,m,n),\\
             \mathcal{P}_{\alpha_{t+1,l,m},\gamma}(x)\cup \mathcal{C}_{l,m}(\gamma),\ else.
             \end{array}
\right.
\end{equation}
if pruning is triggered, we compute the mean value of the feature activations at the current node as the importance metric. According to the corresponding pruning rate, the top‑a tokens are then selected based on the Top‑K operation. These important tokens continue to participate in subsequent computations, while the remaining tokens are replaced with the cached results from previous steps.

We use the feature mean to measure token importance rather than employing the modeled sensitivity error, mainly because the pruning positions are highly correlated with the generated content. As previously discussed, our modeled sensitivity error is primarily model‑specific and largely independent of the generation content.
We use the feature mean instead of the more common attention weights because obtaining attention weights requires explicitly constructing intermediate attention computations, which conflicts with the widely adopted FlashAttention acceleration mechanism.

During pruning, both the pruning timing and pruning rate are fully determined by the offline‑modeled sensitivity error, thus introducing no additional computational overhead during inference. The cost of computing the pruning positions is already included in the reported inference speed of the main experiments.

\subsection{Overall Analysis}
\label{overall}

\textbf{Generalization and Limitation.}
Our sensitivity modeling is analogous to downloading model weights, only a one-time setup is required for permanent use.
When the model is fine-tuned or changed with a new model, SODA requires sensitivity re-modeling. Note that it is fully automated, without manual heuristics, and generalizes across different image sizes, CFGs and seeds on the same model.

\textbf{Scalability.}
SODA's modeling is conducted via random generations, so its time depends on the model inference. Larger models incur longer modeling time. As modeling is insensitive to sample count, we reduce it to mitigate overhead (e.g., 10 for video vs. 100 for image models).

\section{More Experiments}
\label{exper}
In the experiments, we provide the model implementation details omitted in the main text, along with the parameters used for the baselines and the specifications of the benchmarks and evaluation metrics.

\begin{table*}[!ht]
    \centering
    \scalebox{1.0}{
    \setlength{\tabcolsep}{6pt}
    \begin{tabular}{cccccccccc}
    \hline
        Model & Sampling & Latency(s)↓ & FLOPs(G)↓ & Speed↑ & FID↓ & sFID↓ & IS↑ & Precision↑ & Recall↑ \\ \hline
        DiT & DDPM & 2.508 & 118.68 & 1.00× & 2.23 & 4.57 & 275.65 & 0.82 & 0.58 \\
        \textbf{SODA} ($N_s$=125) & DDPM & 1.778 & 76.70 & 1.55× & 2.21 & 4.72 & 272.01 & 0.82 & 0.58 \\
        \textbf{SODA} ($N_s$=72) & DDPM & 1.070 & 43.42 & 2.73× & 2.47 & 5.09 & 262.30 & 0.81 & 0.59 \\
        \textbf{SODA} ($N_s$=69) & DDPM & 0.839 & 35.25 & 3.37× & 2.88 & 6.11 & 252.03 & 0.80 & 0.59 \\
        \textbf{SODA} ($N_s$=63) & DDPM & 0.704 & 29.66 & 4.00× & 3.33 & 6.76 & 237.69 & 0.78 & 0.60 \\ \hline
        DiT & DDIM & 0.533 & 23.74 & 1.00× & 2.25 & 4.33 & 239.97 & 0.80 & 0.59 \\
         \textbf{SODA} ($N_s$=31) & DDIM & 0.433 & 15.61 & 1.52× & 2.37 & 4.51 & 242.22 & 0.82 & 0.59 \\
        \textbf{SODA} ($N_s$=25) & DDIM & 0.314 & 12.21 & 1.94× & 2.39 & 4.55 & 240.75 & 0.81 & 0.58 \\
        \textbf{SODA} ($N_s$=18) & DDIM & 0.263 & 9.55 & 2.49× & 2.75 & 4.56 & 235.65 & 0.80 & 0.58 \\
        \textbf{SODA} ($N_s$=12) & DDIM & 0.162 & 6.25 & 3.80× & 3.62 & 5.88 & 228.41 & 0.79 & 0.59 \\ \hline
    \end{tabular}}
    \caption{\textbf{More results of DiT-XL/2 on class-conditional image generation.} The arrow denotes whether lower or higher values indicate superior performance. $N_s$ represents the number of cache intervals and is used to adjust the acceleration budget. All speeds are measured on RTX 4090 GPU.}
    \label{tab:dit2}
\end{table*}

\subsection{More Implementation Details}
\label{detail}
We apply the SODA model to DiT‑XL/2, PixArt-$\alpha$, and OpenSora, which represent category‑to‑image, text‑to‑image, and text‑to‑video generation tasks, respectively.

\subsubsection{DiT-XL/2}
Experiments conducted on DiT-XL/2~\cite{peebles2023scalable} employ both DDPM~\cite{ho2020denoising} and DDIM~\cite{song2020denoising} samplers. The DDPM and DDIM samplers that we adopt consist of 250 and 50 steps, respectively.

For DDPM, implementation of different methods is as follows: FORA~\cite{selvaraju2024fora} adopts a fixed caching interval of $\mathcal{N}=3$. 
ToCa~\cite{zou2024accelerating} employs an average caching ratio of $R=93\%$ and an average caching interval of $\mathcal{N}=4$, and DuCa~\cite{zou2024accelerating1} employs a caching ratio of $R=95\%$ and an average caching interval of $\mathcal{N}=3$.
For our method SODA, we perform a pre-inference error analysis using 100 images, setting $\lambda = 0.3$ and $\beta = 0.4$. The overall acceleration efficiency is controlled via the number of caching intervals, denoted by $N_s$.

For DDIM, FORA adopts a fixed caching interval of $\mathcal{N}=3$. ToCa employs a caching ratio of $R=93\%$ and an average caching interval of $\mathcal{N}=3$, and DuCa employs an average caching ratio of $R=95\%$ and an average caching interval of $\mathcal{N}=3$. The basic settings of our SODA are kept consistent with those in DDPM.

\subsubsection{PixArt}
For PixArt-$\alpha$~\cite{chen2023pixart}, we adopt the DPM++~\cite{lu2025dpm} with 20 steps as sampler. The implementation of different methods is as follows: FORA adopts a fixed caching interval of $\mathcal{N}=2$. ToCa employs an average caching ratio of $R=70\%$ and an average caching interval of $\mathcal{N}=3$, and DuCa employs an average caching ratio of $R=25\%$ and an average caching interval of $\mathcal{N}=3$. For our method SODA, we perform a pre-inference error analysis using 100 images, setting $\lambda = 0.4$ and $\beta = 0.8$. The overall acceleration efficiency is controlled via the number of caching intervals, denoted by $N_s$.

\subsubsection{OpenSora}
Experiments on OpenSora are conducted with 30 reflective flow steps. The implementation of different methods is as follows:
FORA adopts a fixed caching interval of $\mathcal{N}=2$. ToCa sets different average caching intervals for different modules, with $\mathcal{N}=3$ for temporal attention, spatial attention, and MLP, and $\mathcal{N}=6$ for cross-attention. The average caching ratio is set to $100\%$ for all modules except the MLP, which has an average caching ratio of $85\%$. DuCa adopts the same scheme as ToCa. $\mathrm{\Delta}$-DiT~\cite{chen2024delta} divides the complete denoising process into two stages. PAB~\cite{zhao2024real} sets the caching interval to $\mathcal{N}=2$ for temporal attention, $\mathcal{N}=4$ for spatial attention, $\mathcal{N}=6$ for cross-attention. The MLP module is fully computed.
For our method SODA, we perform a pre-inference error analysis using 10 videos to reduce the time cost of error analysis, setting $\lambda = 0.1$ and $\beta = 0.0$ for higher acceleration. The overall acceleration efficiency is controlled via the number of caching intervals, denoted by $N_s$.

\subsubsection{Evaluation and metrics}
For class-conditional image generation, we use DiT-XL/2 to randomly generate 50k $256 \times 256$ images from ImageNet~\cite{deng2009imagenet}, use FID~\cite{heusel2017gans}, sFID, IS~\cite{salimans2016improved}, Precision, and Recall to measure the quality of the generated images.
The Fréchet Inception Distance (FID) measures the distributional discrepancy between real and generated images in the Inception‑V3 feature space, where a lower score indicates higher fidelity and realism. Its spatial variant, sFID, computes feature distribution differences at the spatial level rather than the global pooling layer, making it more sensitive to structural and texture coherence. The Inception Score (IS) evaluates image quality and diversity based on the confidence and entropy of predicted class probabilities from an Inception network. A higher score implies clearer and more diverse generations. Precision quantifies perceptual fidelity by measuring how many generated samples fall within the manifold of real data, while Recall evaluates diversity by estimating the coverage of the real image distribution. Together, these metrics provide a comprehensive evaluation of the trade‑off between fidelity and diversity in class‑conditional image generation.

For text‑to‑image generation, we evaluate performance using FID‑30K and the CLIP Score~\cite{hessel2021clipscore}, which jointly assess generation quality and text‑image alignment of images generated by PixArt-$\alpha$. FID‑30K is computed between 30,000 generated images and the corresponding real images from the MS‑COCO2017 dataset\cite{lin2014microsoft}, measuring the Fréchet distance between their feature distributions in the Inception‑V3 embedding space. A lower value indicates that the generated images are closer to real ones, reflecting higher visual fidelity and realism. The CLIP Score evaluates semantic consistency between an image and its associated text prompt by computing the cosine similarity of their representations in the CLIP model’s joint vision‑language embedding space. A higher CLIP Score denotes stronger semantic alignment and better text‑image correspondence. Together, these two metrics provide complementary perspectives on both the perceptual quality and the semantic faithfulness of text‑to‑image generation.

For text‑to‑video generation, we adopt VBench~\cite{huang2024vbench} to comprehensively assess the quality of generated videos. We use 950 text prompts and generate five videos per prompt, resulting in a total of 4,750 videos, each rendered at 480p resolution with a 9:16 aspect ratio and 2 seconds duration. VBench evaluates 16 aspects covering semantic alignment, visual quality, temporal consistency, and aesthetic appeal, providing a holistic measurement of video generation performance.

\subsection{More Experiment Results}
\label{results}

To supplement the experimental results presented in the main paper and further demonstrate the effectiveness of our method, this section includes additional acceleration results and ablation studies.

\subsubsection{More results of Dit-XL/2}
We provide additional results of the SODA method under various acceleration ratios in Tab.~\ref{tab:dit2}.

It can be observed that our method preserves most of the model performance under low acceleration ratios. Due to the minimal accumulated cache error obtained via dynamic programming and the adaptive pruning decision that further reduces the cache error, our method can even improve the performance of the original model in certain cases. For example, the results of DDPM with $N_s$=125 and DDIM with $N_s$=31.

Moreover, as the acceleration ratio continues to increase, the model inevitably begins to omit important computations after removing redundant ones, leading to a gradual degradation in performance.

This phenomenon reflects the natural trade-off between generation quality and acceleration. The advantage of our method lies in its ability to improve model performance under low acceleration ratios and significantly delay the trade-off point, thereby preserving more of the model’s performance even as the acceleration increases.

\begin{table}[t]
    \centering
    \scalebox{0.6}{
    \begin{tabular}{cccccccc}
    \hline
        \multirow{2}{*}{Model} & \multirow{2}{*}{Latency(s)↓} & \multirow{2}{*}{FLOPs(G)↓} & \multicolumn{4}{c}{VBench(\%)↑} \\ \cline{4-7}
        & & & Total & \makecell[c]{Imaging\\quality} & \makecell[c]{Subject\\consistency} & Scene \\ \hline
        OpenSora & 102.287 & 3283.20 & 79.13 & \textbf{61.14} & 95.09 & 52.46\\ 
        \rowcolor{gray!20}\textbf{SODA} ($N_s$=21) & \textbf{86.684} & \textbf{2319.94}\textsubscript{\textcolor{red}{1.42×}} & \textbf{79.13} & 60.63 & \textbf{95.26} & \textbf{53.35}\\ \hline
        $\Delta$-DiT (ArXiv24) & - & 3166.47 & 78.21 & - & - & -\\ 
        PAB (ICLR25) & - & 2558.25 & 76.95 & - & - & -\\ 
        FORA (ArXiv24) & - & 1751.32 & 76.91 & - & - & -\\ 
        ToCa (ICLR25) & 54.698 & 1394.03 & 78.34 & 56.04 & 93.92 & 50.21\\ 
        DuCa (ICLR25) & 50.637 & 1315.62 & 78.39 & 56.54 & 94.02 & 50.41\\
        \rowcolor{gray!20}\textbf{SODA} ($N_s$=12) & 49.896 & 1314.42 & \textbf{78.49} & \textbf{58.14} & \textbf{94.60} & \textbf{53.33}\\
        \rowcolor{gray!20}\textbf{SODA} ($N_s$=11) & \textbf{48.022} & \textbf{1277.59}\textsubscript{\textcolor{red}{2.57×}} & 78.41 & 57.35 & 94.14 & 52.00\\ \hline
    \end{tabular}}
    \caption{\textbf{More VBench results of OpenSora on text-to-video generation.} All speeds are measured on 80G A100 GPU.}
    \label{tab:opensora2}
\end{table}

\subsubsection{More results of OpenSora}
We have reported the overall performance on VBench for OpenSora in the main paper. However, since the VBench metric aggregates multiple evaluation criteria with varying importance, the differences between methods on this metric are often minor (typically less than 1), which hinders a clear analysis of our model's advantages.

Considering that VBench primarily focuses on videos depicting motion of specific objects, we argue that analyzing the corresponding aspects of quality, consistency, and overall perceptual experience is more important.

Therefore, we report the relevant detailed metrics within VBench in Tab.~\ref{tab:opensora2} to further highlight the effectiveness of our method.
It can be observed that under a low acceleration ratio (1.42×), our method even outperforms the original model in terms of subject consistency and scene perception, indicating that it is capable of improving performance during acceleration in video generation models.

Moreover, even at the same or higher acceleration ratios compared to the baseline, our method consistently maintains superior performance, with particularly significant advantages in metrics such as image quality.
This indicates that our method can effectively mitigate the error introduced by acceleration, achieving speed-up while preserving generation quality.

\subsubsection{More generation models}
We extend experiments to Qwen-Image~\cite{wu2025qwenimagetechnicalreport} (image) and Wan2.1~\cite{wan2025} (video) to demonstrate our effectiveness.
In Tab.~\ref{tab:more_generation}, ``Base'' is a baseline caching steps with fixed intervals (ToCa/DuCa require manual heuristics, limiting their transferability). We achieve higher fidelity under the same acceleration.

\begin{table}[t]
    % \vspace{-1em}
    \centering
    \resizebox{0.49\textwidth}{!}{
    \setlength{\tabcolsep}{4pt}
    \renewcommand\arraystretch{1.0}
    \begin{tabular}{cccc|ccc}
    \cmidrule(lr){1-4} \cmidrule(lr){5-7}
    Qwen-Img & \makecell[c]{Lat.(s)↓\\/Spe.↑} & CLIP↑ & IR↑ & Wan2.1 & \makecell[c]{Lat.(s)↓\\/Spe.↑} & \makecell[c]{Vbench\\(\%)↑} \\ \cmidrule(lr){1-4} \cmidrule(lr){5-7}
    Origin & 73.431$_{1.00\times}$ & 35.69 & 1.0242 & Origin & 187.46$_{1.00\times}$ &	78.97 \\
    Base & 41.135$_{1.79\times}$ & 35.14 & 0.8677 & Base & 118.82$_{1.58\times}$ & 74.24\\
    \textbf{Ours} & \textbf{40.749}$_{1.80\times}$ & \textbf{35.72} & \textbf{1.0352} & \textbf{Ours} & \textbf{119.46}$_{1.57\times}$ & \textbf{75.61}\\ \cmidrule(lr){1-4} \cmidrule(lr){5-7}
    \end{tabular}}
    % \vspace{-1em}
\caption{\textbf{Results on more generation models.}}
\label{tab:more_generation}
% \vspace{-1em}
\end{table}

\subsubsection{More baselines}
We include more training-free baselines and achieve superior fidelity in Tab.~\ref{tab:more_baseline}  ($\Delta$-DiT is not open-sourced, so we adopt its setting).

\begin{table}[t]
    % \vspace{0.2em}
    \centering
    \resizebox{0.48\textwidth}{!}{
    \setlength{\tabcolsep}{8pt}
    \renewcommand\arraystretch{0.9}
    \begin{tabular}{cccccccc}
    \cmidrule(lr){1-8}
    DiT-XL/2 & Lat.(s)↓ & Spe.↑ & IS↑ & FID↓ & sFID↓ & P↑ & R↑\\ \cmidrule(lr){1-8}
    DDPM/250 & 2.508 & 1.00× & 275.65 & 2.23 & 4.57 & 0.82 & 0.58 \\ 
    $\Delta$-DiT~\cite{chen2024delta} & - & 1.52× & 271.03 & 2.31 & - & - & - \\
    \textbf{Ours} & 1.778 & 1.41× & \textbf{272.01} & \textbf{2.21} & \textbf{4.72} & \textbf{0.82} & 0.58\\
    \textbf{Ours} & \textbf{1.651} & \textbf{1.52×} & 271.77 & 2.23 & 4.76 & \textbf{0.82} & \textbf{0.59}\\ \cmidrule(lr){1-8}
    DDIM/20 & 0.243 & 1.00× & 223.23 & 3.51 & 4.92 & 0.79 & 0.57\\
    GOC(FORA)~\cite{qiu2025accelerating} & 0.167 & 1.46× & 191.44 & 6.78 & 8.74 & 0.74 & 0.53 \\
    % GOC+\textbf{Ours} & \\
    \textbf{Ours} & 0.168 & 1.45× & \textbf{205.05} & \textbf{5.03} & \textbf{5.80} & \textbf{0.76} & \textbf{0.56} \\ \cmidrule(lr){1-8}
    \end{tabular}}
    % \vspace{-1em}
\caption{\textbf{Comparison with more baselines.}}
\label{tab:more_baseline}
% \vspace{-1em}
\end{table}

\subsubsection{Comparison with training methods.}
In Tab.~\ref{tab:more_training}, compared to training caching, SODA achieves the highest IS and outperforms L2C in FID without training.
Compared to distillation, SODA aims for a better balance between acceleration and fidelity, while distillation trades extremely training cost and generation degradation for maximal acceleration.
Overall, SODA's advantage is training-free nature, enabling generalization without manual heuristics.

\begin{table}[t]
    % \vspace{-1em}
    \centering
    \resizebox{0.49\textwidth}{!}{
    \setlength{\tabcolsep}{1pt}
    \renewcommand\arraystretch{1}
    \begin{tabular}{ccccc|ccccc}
    \cmidrule(lr){1-5} \cmidrule(lr){6-9}
    \makecell[c]{Training\\caching} & \makecell[c]{Lat.(s)↓\\/Spe.↑} & IS↑ & FID↓ & sFID↓ & \makecell[c]{Distillation\\one-step} & Spe.↑& \makecell[c]{Fidelity↑\\($\frac{\text{Method}}{\text{Origin}}$)(\%)} & \makecell[c]{Costs↓\\GPU/time} \\ \cmidrule(lr){1-5} \cmidrule(lr){6-9}
    DiT-XL/2 & 0.533$_{1.00\times}$ & 239.97 & 2.25 & 4.33 & DiT-XL/2 & 1× & 100 & - \\
    L2C~\cite{ma2024learning} & None$_{1.25\times}$ & 233.26 & 2.62 & 4.50 & ShortCut~\cite{frans2024one} & $\sim$50× & $\sim$40 & TPUV3s/2days \\
    HarmoniCa~\cite{huang2024harmonica}  & None$_{1.30\times}$ & 238.74 & 2.36 & 4.24 & $\pi$-flow~\cite{chen2025pi} & $\sim$50× & $\sim$78 & Not mentioned \\
    \textbf{Ours} & \textbf{0.403}$_{1.32\times}$ & \textbf{241.71} & 2.37 & 4.58 & \textbf{Ours} & $\sim$4× & $\sim$\textbf{98} & \textbf{Free} \\ \cmidrule(lr){1-5} \cmidrule(lr){6-9}
    \end{tabular}}
    % \vspace{-1em}
\caption{\textbf{Comparison with training-based caching and few-step distillation.}}
\label{tab:more_training}
% \vspace{-1em}
\end{table}

\subsection{More Ablation Study}
\label{ablation}

\begin{table}[t]
    \centering
    \scalebox{0.98}{
    \begin{tabular}{cccccc}
    \hline
        \multirow{2}{*}{Model} & \multicolumn{4}{c}{VBench(\%)↑} \\ \cline{2-5}
        & Total & \makecell[c]{Imaging\\quality} & \makecell[c]{Subject\\consistency} & Scene \\ \hline
        Vanilla     & 76.41 & 54.23 & 93.54 & 50.51 \\
        OFS w/ DCS & 78.37 & 56.54 & 94.12 & 53.41 \\
        OFS w/ UAS & 78.41 & 57.55 & 94.07 & 53.03 \\
        \rowcolor{gray!20} SODA (Ours) & \textbf{78.49} & \textbf{58.14} & \textbf{94.60} & \textbf{53.33}\\ \hline
        
    \end{tabular}}
    \caption{\textbf{Ablation study of OpenSora on text-to-video generation.}}
    \label{tab:ablation2}
\end{table}

\begin{table}[!ht]
    \centering
    \scalebox{0.8}{
    \begin{tabular}{c|c|ccccc}
    \hline
        \multicolumn{2}{c|}{FID↓} & \multicolumn{5}{c}{$\beta$} \\ \cline{3-7}
        \multicolumn{2}{c|}{} & 0.4 & 0.5 & 0.6 & 0.7 & 0.8 \\ \hline
        \multirow{3}{*}{$\lambda$} & 0.4 & 27.46 & 27.44 & 27.42 & 27.38 & 27.33 \\
        ~ & 0.5 & 27.44 & 27.42 & 27.37 & 27.33 & 27.32 \\
        ~ & 0.6 & 27.36 & 27.35 & 27.34 & 27.33 & 27.32 \\ \hline
    \end{tabular}}
    \caption{\textbf{The impact of parameters $\lambda$ and $\beta$ on PixArt-$\alpha$.}}
    \label{tab:params}
\end{table}

\subsubsection{More ablation study on OpenSora.}
To demonstrate the adaptability and generalization ability of our method across different tasks, we not only report additional main experimental results, but also include ablation studies conducted on OpenSora.

As shown in Tab.~\ref{tab:ablation2}, compared to the vanilla caching strategy, our proposed DCS module improves performance by 1.96\% on average by minimizing cumulative caching error through dynamic programming based on estimated error. In addition, the adaptive pruning mechanism (UAS) corrects caching error by retaining important tokens, resulting in a 2\% improvement. The combination of both modules leads to an average increase of 2.08 on the VBench, with nearly a 4\% gain in imaging quality.

On the one hand, this demonstrates the effectiveness of our proposed modules; on the other hand, it highlights the general applicability of our SODA.

\begin{figure*}[t]
\centering
\includegraphics[width=0.98\textwidth]{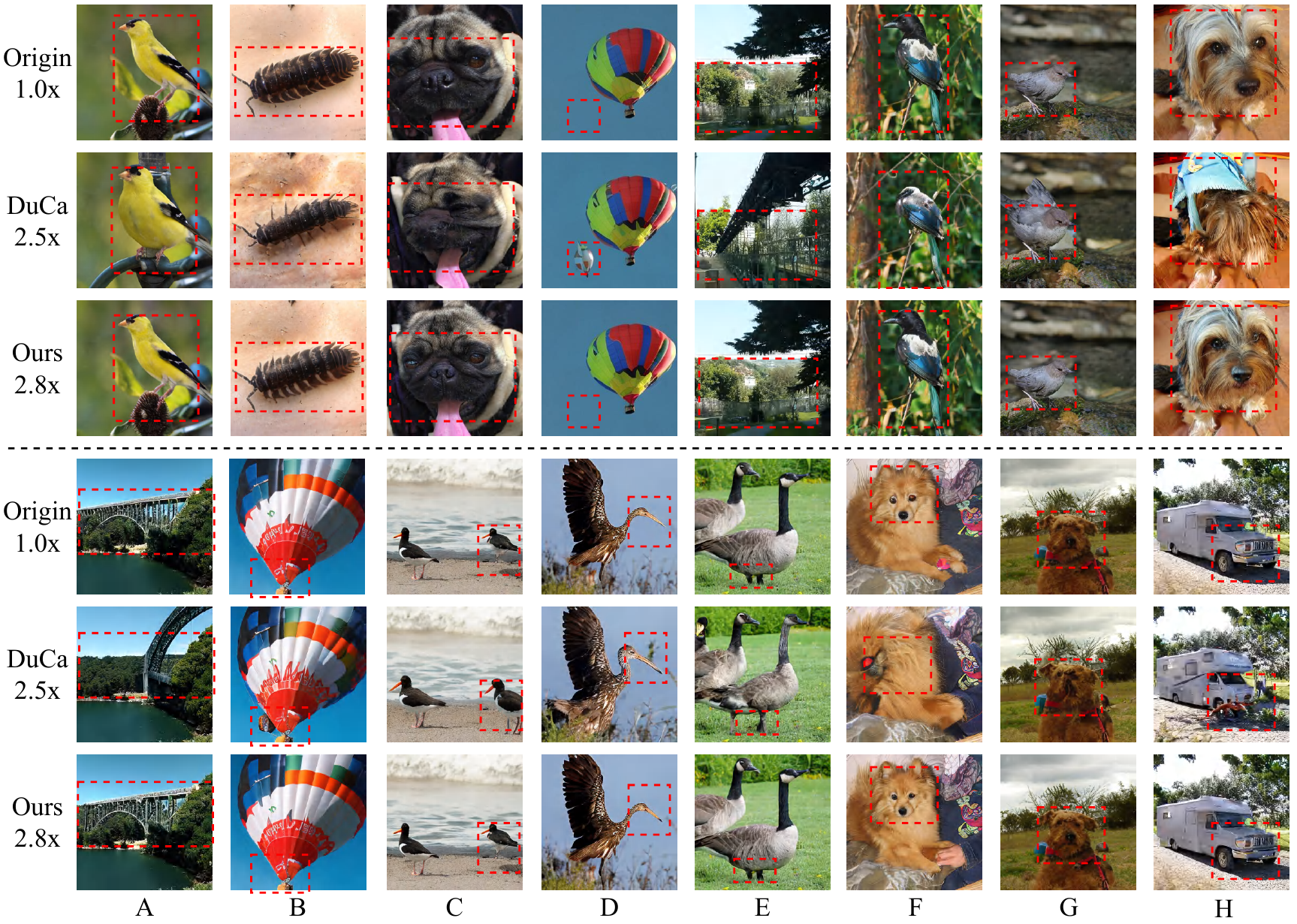} % Reduce the figure size so that it is slightly narrower than the column.
\caption{\textbf{Visualization results of different acceleration methods on DiT-XL/2.} Comparisons are highlighted with red dashed boxes.
}
\label{fig:dit}
\end{figure*}

\begin{figure*}[t]
\centering
\includegraphics[width=0.97\textwidth]{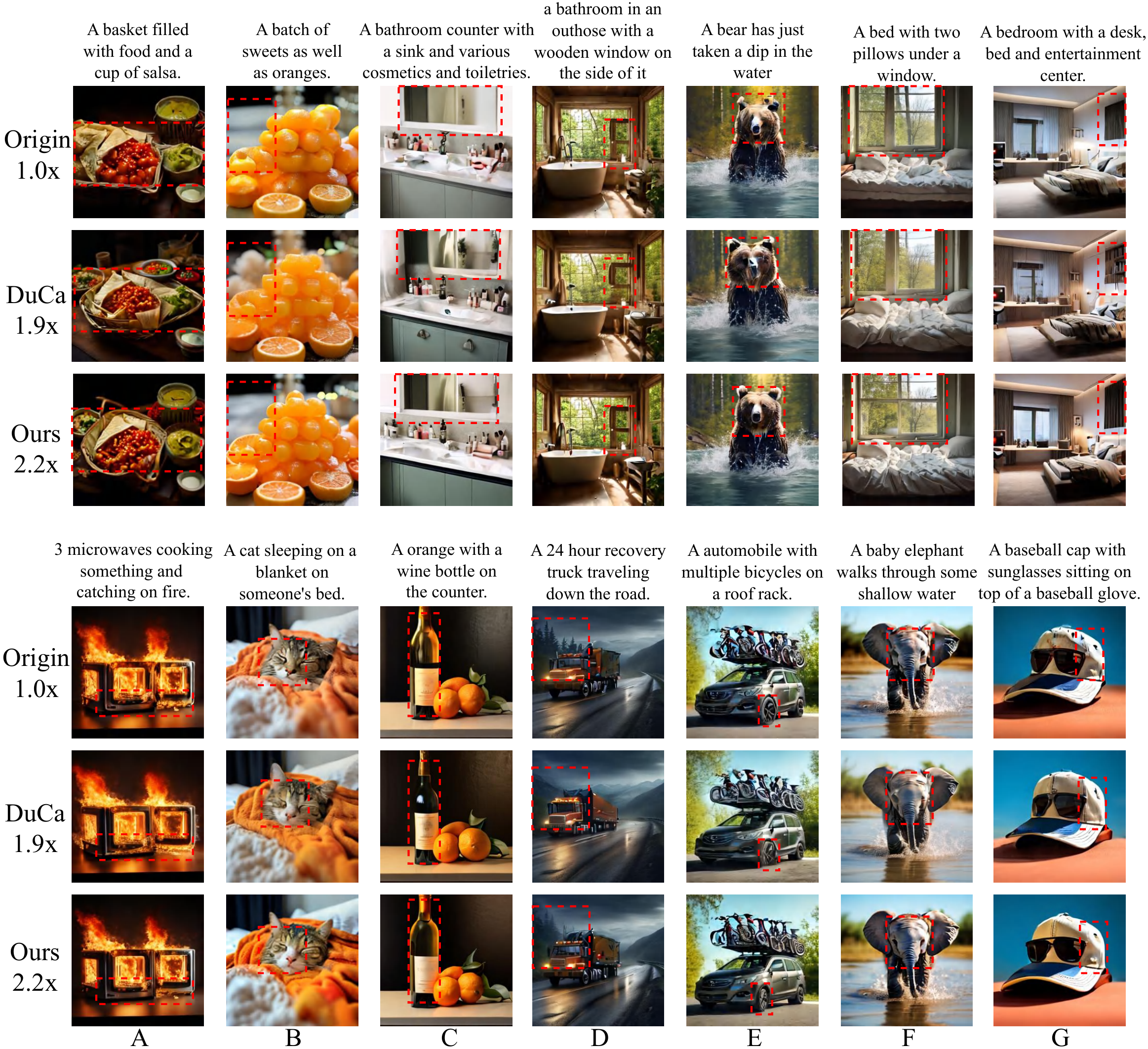} % Reduce the figure size so that it is slightly narrower than the column.
\caption{\textbf{Visualization results of different acceleration methods on PixArt-$\mathrm{\alpha}$}.  
}
\label{fig:pixart}
\end{figure*}

\subsubsection{More parameter analysis on PixArt.}
To further demonstrate the robustness of the hyperparameters introduced by the UAS module, in addition to the parameter space reported on DiT in the main text, we also report results on the parameter space explored on PixArt in Tab.~\ref{tab:params}.

It can be observed that the FID fluctuates within a narrow range of 0.14 across the entire parameter space, indicating a high degree of robustness and suggesting that our method is largely insensitive to the hyperparameters introduced in UAS.

Typically, $\lambda$ is fixed. Given acceleration factor $r>1$, we first obtain the number of cache interval $N_s=N/r-1$ (leave room for pruning), then select $\beta$: $\sum_{t}^{T}\sum_{l}^{L}\{\lambda\mathcal{E}_c(t,l,n)+\beta\}=1/r$ (omit transformation from $t$ to $n$), it means the proportion of pruned tokens is the inverse of speed-up factor. Due to the adaptive pruning decisions, $\beta$ may fluctuate slightly, but it has negligible impact on acceleration factor.

\begin{table}[t]
    % \vspace{-1em}
    \centering
    \resizebox{0.47\textwidth}{!}{
    \setlength{\tabcolsep}{4pt}
    \renewcommand\arraystretch{0.9}
    \begin{tabular}{ccccccccc}
    \cmidrule(lr){1-9}
    DiT-XL/2 & FlashAtt & Lat.(s)↓ & Spe.↑ & IS↑ & FID↓ & sFID↓ & P↑ & R↑ \\ \cmidrule(lr){1-9}
    Mean & \ding{51} & 0.201 & 2.65× & \textbf{233.22} & \textbf{2.80} & \textbf{4.61} & \textbf{0.80} & \textbf{0.58} \\
    Variance & \ding{51} & 0.211 & 2.53× & 233.20 & 2.82 & 2.67 & 0.80 & 0.58 \\
    Norm & \ding{51} & \textbf{0.197} & \textbf{2.70×} & 233.18 & 2.83 & 4.66 & 0.80  & 0.58 \\
    \textcolor{gray!90}{Attn weights} & \textcolor{gray!90}{\ding{55}} & \textcolor{gray!90}{0.300} & \textcolor{gray!90}{1.78×} & \textcolor{gray!90}{233.25} & \textcolor{gray!90}{2.81} & \textcolor{gray!90}{4.63} & \textcolor{gray!90}{0.80} & \textcolor{gray!90}{0.58}
    \\ \cmidrule(lr){1-9}
    \end{tabular}}
    % \vspace{-1em}
\caption{\textbf{Ablation study for pruning location.}}
\label{tab:more_ablation}
% \vspace{-1em}
\end{table}

\subsubsection{More comparison on pruning methods}

In Tab.~\ref{tab:more_ablation}, we use feature mean and top-K selection to determine pruning positions, to identify important tokens with a simple and efficient way, \textbf{avoiding} large time overhead. Attention weights are excluded due to incompatibility with FlashAttention, despite their effectiveness.

\section{More Visualization}
\label{visualization}

We provide additional visual comparisons of acceleration results, along with detailed example analyses.

\subsection{DiT-XL/2}
\label{visual_dit}

It can be observed that our method preserves most of the model performance under low acceleration ratios. Due to the minimal accumulated cache error obtained via dynamic programming and the adaptive pruning decision that further reduces the cache error, our method can even improve the performance of the original model in certain cases. For example, the results of DDPM with $N_s$=125 and DDIM with $N_s$=31.

Moreover, as the acceleration ratio continues to increase, the model inevitably begins to omit important computations after removing redundant ones, leading to a gradual degradation in performance.

This phenomenon reflects the natural trade-off between generation quality and acceleration. The advantage of our method lies in its ability to improve model performance under low acceleration ratios and significantly delay the trade-off point, thereby preserving more of the model’s performance even as the acceleration increases.

\begin{figure*}[t]
\centering
\includegraphics[width=0.98\textwidth]{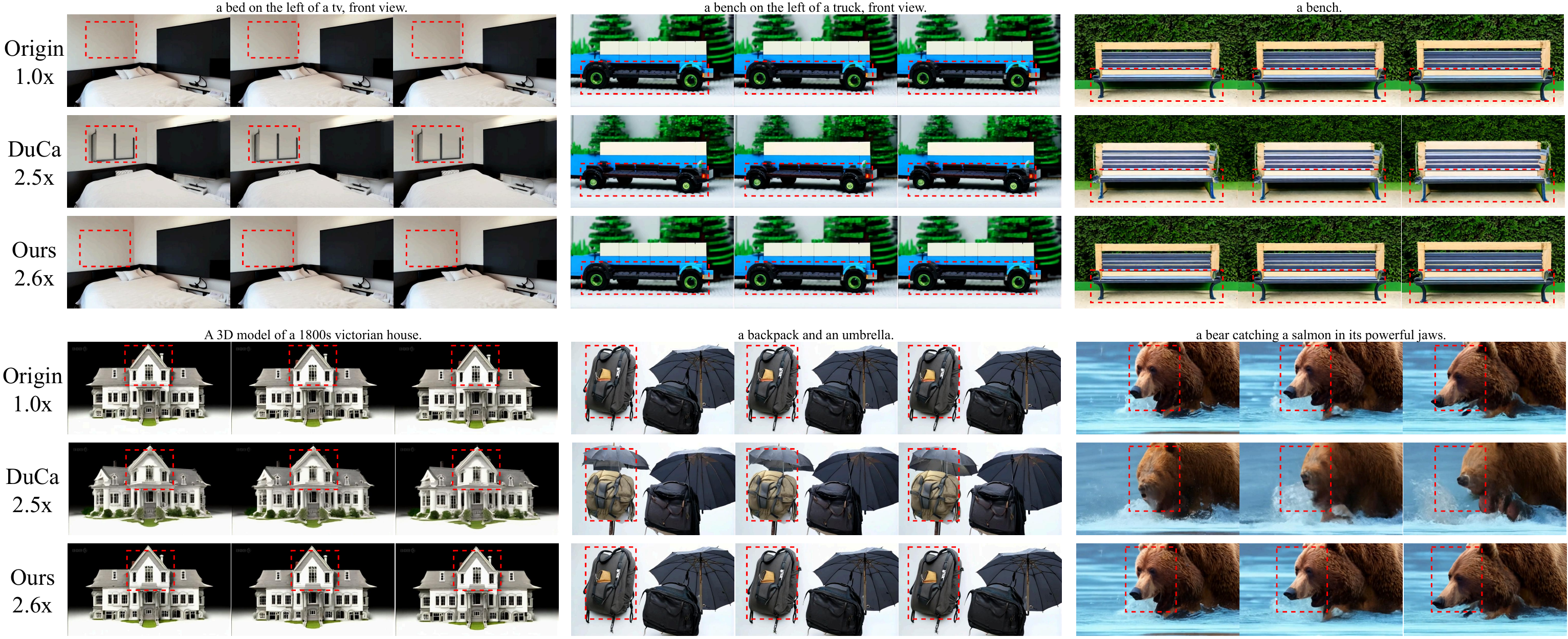} % Reduce the figure size so that it is slightly narrower than the column.
\caption{\textbf{Visualization results of different acceleration methods on OpenSora.}}
\label{fig:opensora}
\end{figure*}

Our method SODA obtains further acceleration, with the generation speed $180\%$ faster than the original model, and $12\%$ faster than DuCa. The visualization results are shown in Fig.~\ref{fig:dit}. Compared with DuCa, SODA-generated images demonstrate finer detail clarity (e.g., the face of the dogs in the upper Col.C and the lower Col.F), while images from DuCa exhibits notable degree of distortion. 

SODA-generated images possess better background texture and overall alignment with the images from the original model (e.g., the background of the woodlouse in upper Col.B and the overall structure of the bridge in lower Col.A). 

SODA also mitigates some of the defects in DuCa-generated images. For example, numeracy confusion (e.g., a third duck by DuCa in lower Col.E, and a second hot-air balloon by DuCa in upper Col.D) and redundant attributes(e.g., the hat on the dog head by DuCa in upper Col.H).
% detail clarity, background texture, overall fidelity, numeracy confusion, distortion, redundant attributes or objects TBD

\subsection{PixArt}
\label{visual_pixart}

Our method SODA achieves a higher acceleration ratio of $2.2\times$ compared with the $1.9\times$ of DuCa. The visualization results are shown in Fig.~\ref{fig:pixart}. SODA-generated images exhibit better detail and overall consistency with those generated by the original model (e.g., the basket of food in upper Col.A and the details of the recovery truck and the bear in lower Col.C and upper Col.D, respectively). 

It should be noted that our method sometimes even demonstrate superior generation quality compared with the original model (e.g., the right ear of the bear in upper Col.E is more complete, and the light shade on the wine bottle in lower Col.C is more real). This aligns with our quantitative analysis in the Experiment Section.

\subsection{OpenSora}
\label{visual_opensora}

Our method SODA achieves an extra $0.1\times$ acceleration based on DuCa, which is now $160\%$ faster than the original model. The visualization results are shown in Fig.~\ref{fig:opensora}. Compared with DuCa, our method mitigates problems such as object redundancy (e.g., the redundant decoration above the bed by DuCa in the upper left frames) and detail distortion (e.g., the distorted face of the bear by DuCa in the lower right frames). 

Our method can also better maintain the consistency with the original model (e.g., in the lower middle frames, DuCa-generated frames contain two umbrellas and the color of the backpack on the left is khaki, while there is only one umbrella and the color of the backpack on the left is gray in the frames generated by both the original model and our method). 

Moreover, our method can better handle the relationship of foreground objects and background. For example, in the upper right frames, there exists a clear boundary between the foreground bench and the background bushes in the frames generated by our method. However, in the frames generated by DuCa, the edges of the bench already blend slightly with the bushes.

As illustrated by the visual examples, SODA effectively reduces the caching error while achieving acceleration.

\section{Related Work}
\label{related}

\subsection{Diffusion Transformer}
\label{dit}

\subsubsection{Diffusion Model.}
Diffusion models~\cite{rombach2022high,ho2020denoising} synthesize visual content from pure noise by progressively denoising it through noise prediction. The training and inference process of diffusion models can be divided into two stages: noise diffusion stage and reverse denoising stage, respectively.

During the noise diffusion stage, noise of different magnitudes is added to the original image from the training set with respect to the timestep $t \in [1, T]$, where $T$ is the total number of steps. The noise diffusion process at timestep $t$ can be expressed as:

\begin{equation}
    \boldsymbol{x_{t}} = \sqrt{\bar a_{t}}\boldsymbol{x_{0}} + {\sqrt{1- \bar a_{t}}}\boldsymbol{\epsilon},
\end{equation}
where $\boldsymbol{x_{t}}$ is the noisy image at timestep $t$, $\boldsymbol{x_{0}}$ is the original image, $a_{t}$ is a constant related to $t$ that controls the noise magnitude, and $\boldsymbol{\epsilon} \sim \mathcal{N}(\boldsymbol{0}, \boldsymbol{I})$ is the random noise sampled from the Gaussian distribution.

The model then learns to predict the noise added at a certain timestep by minimizing the following objective~\cite{ho2020denoising}:

\begin{equation}
    \mathcal{L}(\boldsymbol{\theta})=\mathbb{E}_{t}[||\boldsymbol{\epsilon}-\boldsymbol{\epsilon_{\theta}}(\sqrt{\bar a_{t}}\boldsymbol{x_{0}} + {\sqrt{1- \bar a_{t}}}\boldsymbol{\epsilon}, t)||^{2}],
\end{equation}
where $\boldsymbol{x_0}\sim q(\boldsymbol{x}),\boldsymbol{\epsilon} \sim \mathcal{N}(\boldsymbol{0}, \boldsymbol{I})$, $q(\boldsymbol{x})$ is the distribution of the training data, and $\boldsymbol{\epsilon_{\theta}}$ is the estimated noise with the noisy image $\boldsymbol{x_{t}}$ and timestep $t$ as input, parameterized by $\boldsymbol{\theta}$.

During the reverse denoising stage, the trained model predicts the noise added at a certain timestep and removes it to obtain a less noisy image. When reversing from timestep $T$ to $1$, random noise can be gradually denoised to a specific image. The specific denoising process varies among different solvers. Taking DDPM~\cite{ho2020denoising} as an example, the denoising process from $\boldsymbol{x_t}$ to $\boldsymbol{x_{t-1}}$ can be modeled as:
\begin{equation}
    \boldsymbol{x_{t-1}} = \frac{1}{\sqrt{a_{t}}}(\boldsymbol{x_{t}} - \frac{1-a_{t}}{\sqrt{1-\bar a_{t}}}\boldsymbol{\boldsymbol{\epsilon}_{\theta}}(\boldsymbol{x_{t}},t))+\sigma_{t}\boldsymbol{z},
\end{equation}
where $a_{t}$, $\bar a_{t}$, $\sigma_{t}$ are constants related to $t$, and $\boldsymbol{z} \sim \mathcal{N}(\boldsymbol{0}, \boldsymbol{I})$.

\subsubsection{Diffusion Transformer.}
Diffusion transformer~\cite{peebles2023scalable} integrates the transformer~\cite{vaswani2017attention} architecture into the diffusion process, enabling improved controllability and higher generation quality compared with U-Net~\cite{ronneberger2015u}. It has a hierarchical structure consisting of a total number of $L$ DiT blocks. The architecture of DiT and within a specific DiT block $l$ can be expressed as follows:

\begin{equation}
\begin{split}
    &\mathcal{G}=g_{1} \circ g_{2} \circ \cdots \circ g_{L},\\
    &g_{l}=F_{\mathrm{SA}}^{l} \circ F_{\mathrm{CA}}^{l} \circ F_{\mathrm{MLP}}^{l}, 
\end{split}
\end{equation}

where $\mathcal{G}$ denotes the DiT model, $g_{l}$ denotes the DiT block, SA denotes the self-attention layer, CA denotes the cross-attention layer, and MLP denotes the multilayer perceptron. 

Like standard Transformers, the input image of size $H \times W$ is segmented into a number of spatial patches of size $p \times p$, resulting in a total of $\frac{H}{p} \times \frac{W}{p}$ patches. The input $\boldsymbol{x}$ to DiT is a sequence of tokens corresponding to the patches. When $\boldsymbol{x}$ enters a DiT block, it will go through a residual connection, expressed as follows:

\begin{equation}
    f^{l}(\boldsymbol{x})=\boldsymbol{x}+\mathrm{AdaLN} \circ F^{l}_{\star}(\boldsymbol{x}),
\end{equation}
where $l$ denotes the index of DiT block, AdaLN denotes the adapative layer normalization and $\star$ $\in$ \{SA, CA, MLP\}.

Although DiT has achieved substantial progress, its slow inference process remains a major bottleneck, limiting its broader adoption and real-world applicability. The computational overhead largely comes from the attention layer, which has a time complexity of $\mathbf{O}(\boldsymbol{N^2})$, where $\boldsymbol{N}$ is the number of input tokens.

\subsection{Training-free DiT acceleration}
\label{acceleration}
In this paper, we investigate training‑free acceleration approaches that do not require additional training or distillation, thereby avoiding the significant computational overhead typically associated with post‑training optimization. Existing training‑free acceleration methods for diffusion transformers can generally be categorized into two paradigms: feature caching, which reuses intermediate activations for efficiency, and token pruning, a strategy adapted from token reduction techniques originally developed for large language models (LLMs).

\subsubsection{Caching-based Acceleration}
Recently, feature caching has emerged as a training-free method that reuses the results from previous timesteps or layers to avoid redundant computation. Caching-based methods can be categorized into two types according to their granularity: layer-wise caching and token-wise caching.

\textbf{1) Layer-wise caching} can refer to caching the layers in the UNet~\cite{ronneberger2015u} or DiT~\cite{peebles2023scalable} backbones or caching the specific attention or MLP modules. In terms of UNet-based diffusion models,  DeepCache~\cite{ma2024deepcache} and FasterDiffusion~\cite{li2023faster} cache intermediate feature output of certain blocks for later reuse, while TGATE~\cite{liu2024faster} splits the denoising process into two phases and caches the self-attention and cross-attention output, respectively. For diffusion transformer, FORA~\cite{selvaraju2024fora} uses an intuitive caching schedule by computing and caching the output of attention layers and MLP layers in DiT every $\mathcal{N}$ steps, and reusing the results in the subsequent $\mathcal{N}-1$ steps. $\Delta$-DiT~\cite{chen2024delta} caches the deviation between adjacent feature maps to better preserve previous sampling information. TaylorSeer~\cite{liu2025reusing} utilizes the Taylor expansion to optimize approximations in long-range feature reuse scenarios. TeaCache~\cite{liu2025timestep} and AdaCache~\cite{kahatapitiya2024adaptive} adopt adaptive schedules to decide whether to cache or not at a specific timestep.

\textbf{2) Token-wise caching} caches features at a finer granularity. Instead of caching the entire output feature map, it selectively caches some of the tokens based on certain importance or redundancy criteria. ToCa~\cite{zou2024accelerating} selects suitable tokens to cache under the guidance of four scores that comprehensively indicate the influence of caching a specific token. DuCa~\cite{zou2024accelerating1} applies layer-wise and token-wise caching alternately during the denoising process to better balance efficiency and quality.

It should be noted that most of the methods mentioned above adopt a fixed caching schedule (e.g., FORA~\cite{selvaraju2024fora}) or a manually crafted heuristic schedule (e.g., DuCa~\cite{zou2024accelerating1}). Although these methods are intuitive and easy to implement, they are at risk of neglecting the dynamic magnitudes of feature evolvement in different timesteps, layers and modules, and may result in quality degradation. Therefore, current research calls for adaptive methods that can dynamically shift their caching schedules as the denoising process progresses.

\subsubsection{Pruning-based Acceleration}
Token reduction is originally intended for the acceleration of large language models(LLM)~\cite{liu2023tcra,han2024token} and multimodal large language models(MLLM)~\cite{shang2024llava,zhang2024sparsevlm,chen2024image}. However, it also excels in accelerating diffusion models. Token reduction methods can be roughly classified into two classes: token pruning and token merging.

\textbf{1) Token pruning} preserves only a fraction of the total tokens that are the most important and discards the rest before performing compute-intensive operations like attention. Tokens are selected to be preserved or discarded following different criteria. AT-EDM~\cite{wang2024attention} proposes a graph-based algorithm that uses attention maps to measure the importance of tokens. CAT-Pruning~\cite{cheng2025cat} selects tokens based on noise variations across timesteps, selection frequency, and spatial structure. DaTo~\cite{zhang2024token} first identifies tokens with minimal temporal noise difference as base tokens within each patch of the image, and then discards the tokens that have the highest similarity to the base tokens. After the preserved tokens have gone through the intensive computation, the pruned tokens need to be recovered for subsequent computations like convolution. This can be achieved by similarity-based copying or token-wise caching, which is elaborated in the previous subsection.

\textbf{2) Token merging}~\cite{sun2024asymrnr,saghatchian2025cached,bolya2023token} follows similar principles to token pruning. However, it additionally merges the pruned tokens with the preserved ones before omitting them, so as to maintain more information. ToFu\cite{kim2024token} combines both token pruning and merging to achieve optimal performance.

Although token reduction methods demonstrate excellent performance because of their fine-grained operation, intricate pruning strategies are likely to add additional workload, making their efficiency still lag behind caching-based acceleration. In addition, some of these methods leverage the attention maps, which are incompatible with CUDA-based acceleration such as FlashAttention~\cite{dao2022flashattention}.

\subsection{Other Acceleration}
\label{others}
The high latency during the inference stage of diffusion models has drawn continuous attention from the research community, and researchers are working to improve the inference speed of diffusion models from different perspectives.

In general, classic acceleration methods can be categorized into the following domains: \textit{Distillation}~\cite{salimans2022progressive,meng2023distillation, yin2024one} often includes a lightweight student network that derives from the original teacher network through knowledge transfer~\cite{pan2009survey}. \textit{Quantization}~\cite{shang2023post,he2023ptqd,li2023q} transforms the precision of model weights to smaller bit width(\textit{e.g.} FP32 to INT8) via retraining,  thus achieves model compression. Advanced \textit{samplers}, such as DDPM~\cite{ho2020denoising}, DDIM~\cite{song2020denoising}, DPM-Solver~\cite{lu2022dpm}, aim to minimize the sampling steps required to generate an image. However, although these methods achieve an excellent acceleration ratio, most of them demand heavy retraining on high-end GPUs, which is formidable to common practices.

Meanwhile, the acceleration of diffusion models can also be achieved via \textit{hardware-level} optimization~\cite{dao2022flashattention,chen2023speed,tang2025diff,wang2025sd}. 
These methods focus on minimizing memory access frequency or optimizing the efficiency of computing kernels to fully leverage the computing capacity of modern GPUs. 
However, hardware-level optimization often suffers from poor portability, high engineering complexity, and limited adaptability to evolving model architectures.


\end{document}